%% file: main.tex
\documentclass{article} 
\usepackage{iclr2027_conference,times}

\iclrfinalcopy

\usepackage{hyperref}
\usepackage{url}
\usepackage{amsthm}
\usepackage{amsmath,amssymb}
\input{tikz_dependencies}

\usepackage{booktabs}
\usepackage{graphicx}

\newcommand{\E}{\mathbb{E}}
\newcommand{\BE}{\mathbb{E}}
\newcommand{\BZ}{\mathbb{Z}}

\newcommand{\BR}{\mathbb{R}}

\newcommand{\h}{\mathbf{h}}
\newcommand{\bc}{\mathbf{c}}

\newcommand{\M}{\mathcal{M}}

\newtheorem{thm}{Theorem}
\newtheorem*{thm_nonum}{Theorem}

\theoremstyle{remark}
\newtheorem{remark}{Remark}

\graphicspath{{figures/}}

\title{How to track your tangent space: \\ the geometry of the linear representation \\ hypothesis}

\author{Wyatt Mackey\thanks{ Correspondence to \texttt{wyattmackey@gmail.com}. \\  \hspace*{1.53em} Code for our experiments can be found at \href{https://github.com/mackeynations/tangent-lenses/}{https://github.com/mackeynations/tangent-lenses/}},$^{\, \,1,2}$ B. Christopher Rinderspacher,$^1$ Piotr Franaszczuk,$^{1,3}$ \& David L. Boothe.$^1$ \\
\, \\
$^1$DEVCOM Army Research Lab, Aberdeen, MD\\
$^2$Johns Hopkins University, Department of Electrical Engineering, Baltimore, MD\\
$^3$Johns Hopkins University, School of Medicine, Department of Neurology, Baltimore, MD
}

\begin{document}

\maketitle

\pagestyle{fancy}
\fancyhead{} 
\lhead{}

\begin{abstract}
We examine how local geometry shapes interpretability by studying the tangent structure of the activation manifold of language models: we show that hidden states can be decomposed into linear and multidimensional components by projection onto the tangent space. Using this decomposition, we then show that linear components dominate the interpretable signature of the network. We then perform a similar decomposition of the recently proposed $J$-lens, to concentrate its on-manifold versus off-manifold data. We give a full analytic description of a tangent-native $J$-lens, and show that the normal contribution to the $J$-lens depends on the off-manifold extension of the network: in particular, the workspace readout can be changed arbitrarily subject to mild technical constraints, which could prevent a misaligned network from being detected. Empirically, we show that there is not a simple fix for this: the on-manifold Jacobian shows substantial cancellation in mid-to-late layers, and the $J$-lens outputs materially rely on their off-manifold projection.
\end{abstract}

\input{main_body}

\subsection*{AI use statement}

While preparing this work we used generative AI tools for code generation, bug fixes, verifying, correcting, and improving mathematical arguments, including proposing the current form of Theorem \ref{thm:normal_jacobians}, assistance in writing proofs, to verify our data interpretations, to search for related work, to create Figures \ref{fig:lrh_example} and \ref{fig:normal_attack}, and for final proofreading. We verified all code, and wrote all final arguments, including both mathematical claims and the text of the paper, and we verified the relationship of all related work suggested.
We have not used generative AI tools to propose hypotheses or research methodologies, for translation, or for synthetic dataset creation. We take responsibility for the final content of this work,
including text, claims or artifacts produced with the aid of generative AI.

\subsection*{Ethics statement}
This paper discusses methods of evading gradient-based interpretability methods (Theorem \ref{thm:normal_jacobians}), which has potential for misuse. We believe that it is nevertheless in the interest of the field to publish these results: our methods are not difficult to re-derive, and it is important for model auditors to be aware of the potential pitfalls in interpretability tools.

\subsection*{Reproducibility statement}

To ensure reproducibility of our work, we have attached a code base, including scripts to reproduce all experiments described in the paper. In the non-anonymized release, this is published as a public git repository. We have also fully described hyperparameters of our experiments in Appendix \ref{sec:benchmark_and_experiment_details}, and detailed the precise mathematical interventions we perform, so they can be independently recreated without our repository. Likewise, proofs of all mathematical claims are given in Appendix \ref{sec:proofs}. 
%
%
%

\bibliography{iclr2027_conference}
\bibliographystyle{iclr2027_conference}

\appendix
\input{appendix}

\end{document}

%% file: tikz_dependencies.tex
\usepackage[T1]{fontenc}
\usepackage{amsmath,amssymb}
\usepackage{tikz}
\usetikzlibrary{calc,positioning,arrows.meta}

\definecolor{inkdark}{RGB}{15, 23, 42}
\definecolor{slatetext}{RGB}{71, 85, 105}
\definecolor{slatemuted}{RGB}{100, 116, 139}
\definecolor{bordergray}{RGB}{203, 213, 225}
\definecolor{cardbg}{RGB}{248, 250, 252}
\definecolor{trackbg}{RGB}{241, 245, 249}

\definecolor{basedark}{RGB}{30, 58, 138}
\definecolor{baseprimary}{RGB}{29, 78, 216}
\definecolor{basebar}{RGB}{37, 99, 235}
\definecolor{basemed}{RGB}{147, 197, 253}
\definecolor{baselight}{RGB}{239, 246, 255}
\definecolor{basepill}{RGB}{219, 234, 254}

\definecolor{moddark}{RGB}{146, 64, 14}
\definecolor{modprimary}{RGB}{180, 83, 9}
\definecolor{modbar}{RGB}{217, 119, 6}
\definecolor{modmed}{RGB}{252, 211, 77}
\definecolor{modlight}{RGB}{255, 251, 235}
\definecolor{modpill}{RGB}{254, 243, 199}

\definecolor{demoteddark}{RGB}{190, 18, 60}
\definecolor{demotedprimary}{RGB}{225, 29, 72}
\definecolor{demotedlight}{RGB}{255, 241, 242}
\definecolor{demotedtrack}{RGB}{255, 228, 230}

\newcommand{\jlensrow}[9]{%
  \ifnum#9=1
    \node[anchor=west, font=\sffamily\bfseries\footnotesize, text=#8] at (#1+0.16, #2) {\##3};
    \node[anchor=west, font=\ttfamily\bfseries\footnotesize, text=inkdark] at (#1+0.78, #2) {#4};
  \else
    \node[anchor=west, font=\sffamily\footnotesize, text=slatetext] at (#1+0.16, #2) {\##3};
    \node[anchor=west, font=\ttfamily\footnotesize, text=inkdark] at (#1+0.78, #2) {#4};
  \fi
  \fill[fill=trackbg, rounded corners=1.2pt]
    (#1+1.80, #2-0.052) rectangle (#1+2.44, #2+0.052);
  \pgfmathsetmacro{\barw}{max(0.025, 0.64 * #6)}
  \fill[fill=#7, rounded corners=1.2pt]
    (#1+1.80, #2-0.052) rectangle (#1+1.80+\barw, #2+0.052);
  \ifnum#9=1
    \node[anchor=west, font=\sffamily\bfseries\scriptsize, text=inkdark] at (#1+2.50, #2) {#5};
  \else
    \node[anchor=west, font=\sffamily\scriptsize, text=slatetext] at (#1+2.50, #2) {#5};
  \fi
}

\newcommand{\drawminimalrobot}[6]{%
  \draw[#3, line width=1.2pt, line cap=round] (#1, #2+0.20) -- (#1, #2+0.31);
  \filldraw[fill=#5, draw=#3, line width=1.0pt] (#1, #2+0.355) circle (0.045cm);
  \filldraw[fill=#4, draw=#3, line width=1.0pt, rounded corners=1pt]
    (#1-0.305, #2-0.08) rectangle (#1-0.25, #2+0.08);
  \filldraw[fill=#4, draw=#3, line width=1.0pt, rounded corners=1pt]
    (#1+0.25, #2-0.08) rectangle (#1+0.305, #2+0.08);
  \filldraw[fill=#4, draw=#3, line width=1.3pt, rounded corners=4.5pt]
    (#1-0.25, #2-0.20) rectangle (#1+0.25, #2+0.20);
  \fill[fill=#3] (#1-0.115, #2+0.055) circle (0.038cm);
  \fill[fill=#3] (#1+0.115, #2+0.055) circle (0.038cm);
  \draw[#3, line width=1.1pt, line cap=round] (#1-0.11, #2-0.085) -- (#1+0.11, #2-0.085);
  \draw[#3, line width=0.8pt] (#1-0.066, #2-0.115) -- (#1-0.066, #2-0.055);
  \draw[#3, line width=0.8pt] (#1,       #2-0.115) -- (#1,       #2-0.055);
  \draw[#3, line width=0.8pt] (#1+0.066, #2-0.115) -- (#1+0.066, #2-0.055);
  \ifnum#6=1
    \filldraw[fill=modpill, draw=modprimary, line width=0.8pt, rounded corners=2pt]
      (#1+0.06, #2-0.275) rectangle (#1+0.48, #2-0.12);
    \node[font=\sffamily\bfseries\tiny, text=moddark] at (#1+0.27, #2-0.197) {+LoRA};
  \fi
}

%% file: main_body.tex
\section{Introduction}
Monitoring and safeguarding modern language models relies on the development of powerful interpretability tools, designed to give us an understanding of both the structural mechanisms and the underlying thought processes of these models. The latter has been the subject of a wide array of developments in recent years, including sparse autoencoders (SAEs) \citep{huben2024sparse}, logit lenses \citep{nostalgebraist2020logitlens}, and linear classifiers. All of these approaches share a basic underlying structure, in which a linear map 
transforms hidden state values into a more interpretable structure, differing primarily in their training signal.

Underlying these linear decompositions is the \emph{linear representation hypothesis} \citep{elhage2022toy}. The linear representation hypothesis posits that network latents can be understood as a sparse linear combination of an underlying representative dictionary. This implies a specific geometric structure of activation space, as a conic, stratified manifold. We observe that there is a simple test for this underlying geometric claim: any latent $h_\ell \in \M_\ell$ will lie in its own tangent space $T_{\h_\ell}\M_\ell$. This is not the case for generic manifolds. In fact, we show that the projection $P_{T_{\h_\ell}\M_\ell}\h_\ell$ decomposes $\h_\ell$ into the approximately linear component of $\h_\ell$ and its nonlinear data. We show on a simple grokking task that these projections cleanly and causally separate curved and conic components. We then show across a synthetic evaluation suite from \citep{gurnee2026jlens} that the interpretable content of the latent seems to be concentrated on the tangent space projection, with notable exceptions in synthetic tasks with known curved geometry. 

To evaluate this, we use a recent improvement to the classical logit lens \citep{nostalgebraist2020logitlens}, \citet{gurnee2026jlens}'s $J$-lens. The $J$-lens observes that though the unembedding map $U$ is a powerful decoder on late model layers, it is not well adapted to early-to-mid layer structures in the model. In order to align the spaces, the $J$-lens calculates the expected Jacobian of the map from layer $\ell$ to the final layer $L$ of the model, as a strong linear approximation to the downstream causal effect of layers $\ell+1$ through $L$. The composite map $U \circ \BE[J_{\ell \to L}]$ then gives layer-adapted token probabilities, reflecting the downstream aggregate impact of a hidden state $h_\ell$. 

The Jacobian in \citep{gurnee2026jlens} is defined on the ambient space $\BR^d$, but the activation space of the network is much smaller \citep{ansuini2019intrinsic, valeriani2023geometry}. When localized to the activation manifold, the Jacobian of a map of smooth manifolds $M \to N$ is a bundle map $TM \to TN$, and defines a natural on-manifold $J$-lens by projecting away normal components of the Jacobian.

This is much easier to describe mathematically than the $J$-lens of \citet{gurnee2026jlens}: in Theorem \ref{thm:LRH_tangent_jacobians}, we show that when the activation manifold is defined by the linear representation hypothesis, the tangent-projected $J$-lens is a linear map which sends each primitive of the space to the weighted average of outputs when the primitive is present, minus the average output when the primitive is absent. Importantly, we then show that the full $J$-lens is not determined by the network's activations on the data manifold. Unlike SAEs, linear probes, the logit-lens, or the tangent-restricted $J$-lens, the full $J$-lens can be arbitrarily modified along directions not everywhere-tangent to the activation manifold without changing the network's underlying empirical distribution (Theorem \ref{thm:normal_jacobians}). We train a small LoRA adjuster to demonstrate this vulnerability, erasing concepts from the $J$-lens readout without changing the model's outputs. 

Despite this vulnerability, our initial experiments showed that the manifold-native $J$-lens performed substantially worse than a variety of controls: empirically, sufficient cancellation takes place to remove much of the signal from the tangent Jacobian in mid-to-late layers. This suggests that some of the success reported in \citet{gurnee2026jlens} depends materially on the specific off-manifold extension implemented by the transformer. While the off manifold extension is not arbitrary---it is shaped by implicit and explicit regularization, as well as the capcity of the network to realize the on-manifold function---it is difficult to make guarantees about the information that it preserves. 

In summary, our contributions are:
\begin{itemize}
    \item We introduce tangent projection as an intrinsic, local, geometric tool for decomposing language model latents into primarily linear and primarily non-linear components. We then show that linear components of this decomposition control the interpretable directions in the network: \textbf{directions which are linearly extensible \emph{and} on-manifold contain most of the interpretable signature measured by the $J$-lens. }
    \item We provide a theoretical analysis of the $J$-lens, showing that it decomposes into a term that depends on the extension of the model in off-manifold directions, and a term that depends on its on-manifold values and the curvature of the activation manifold. {\bf We show that the $J$-lens can be freely modified, subject to hypotheses, without changing the network's output distribution by perturbing its smooth off-manifold extension.}
    \item We then show empirically that the performance of the $J$-lens is driven by its normal extension. In particular, {\bf the tangent projected constructions of the $J$-lens we test substantially reduce performance, and we empirically demonstrate that $J$-lens information can be erased by normal bundle based attacks}.
\end{itemize}

\section{Related work}

\noindent {\bf Activation manifolds.} Many past authors have noted the low dimensional nature of activation manifolds \citep{ansuini2019intrinsic, valeriani2023geometry}, and the characteristic dimension shifts between network layers. \citet{wurgaft2026manifold} shows that steering network activations along these manifolds smoothly interpolates model choices, while off-manifold steering can destroy model fluency. Similarly to our analysis of the linear representation hypothesis, \citet{robinson2024structure} show that the token space of language models has the structure of a stratified manifold. More recently, authors have explored nonlinear components of the activation manifold \citep{engels2025not}, and proposed mixed decompositions, such as Minkowski sums of low-dimensional manifolds \citep{bhalla2026sparseautoencoderscaptureconcept}. 

\noindent {\bf Geometry in interpretability.} \citet{zaher2026geometric} shows that SAE reconstruction quality depends on a number of intrinsic geometric factors, including curvature. \citet{fel2026structuring} modifies the sparsity structure of SAEs to allow them to capture multidimensional features in natural ways. Sequence-aware SAEs \citep{lubana2026priors, bhalla2026temporal} implicitly approximate tangent and normal components by projecting onto the span of preceding tokens, and passing only the normal component of the hidden state to the SAE.

\noindent {\bf Tangent \& normal constructions.}
Gradient attributions have previously been observed to be strongest when manifold aligned \citep{bordt2023manifold}. Most similar to us, \citet{anders2020fairwashing} shows that gradient attribution on classifiers can be manipulated by modifying the embedding of the normal space of the data manifold---this is very similar to our Theorem \ref{thm:normal_jacobians}, which applies a very similar argument for activation manifolds. 

\section{Background}
\subsection{Notation}
We evaluate transformer architecture language models \citep{vaswani2017attention}, which alternate between applying attention and perceptron layers. The transformer combines the outputs in a residual stream \citep{elhage2021mathematical}. The residual stream is a sequence of vector spaces $H_\ell \cong \BR^d$, where $\ell \in \{0, 1, ..., L\}$ denotes the layer index. The hidden state of a transformer at layer $\ell$ and token $t$ is a vector $\h_{\ell,t} \in H_\ell$. When $t$ can be inferred from context, we simply write $\h_\ell \in H_\ell$. Given a subspace $V \subset H_\ell$, we write $P_V$ for the orthogonal projection onto $V$. We write $T_x\M$ for the tangent space at $x$ in $\M$. When it is clear from context, we will abbreviate this by $T_x$. We will also often write $P_T$ to refer to the projection $P_{T_x\M}$, and similarly for the normal bundle $N\M$.

\subsection{The $J$-lens}
The logit lens was one of the first language-model specific interpretability tools introduced. It functions by applying the unembedding map of the network $U$ early: the idea is that the network needs to decode its representations into the basis of $U$ at the final layer, and the easiest way to do this is to organize concepts in that basis throughout processing---a similar idea motivates cross-layer transcoders \citep{lindsey2024sparse}. \citet{gurnee2026jlens} proposed using the average Jacobian $\BE[\frac{\partial \h_{L, t+k}}{\partial \h_{\ell,t}}]$ as a first order approximation to the map of hidden spaces $H_\ell \to H_L$, emphasizing causality of the hidden states on the outputs. The Jacobian naturally decomposes into on-manifold directions and normal directions: we explore how this local geometry shapes the $J$-lens in Section \ref{sec:decomposing_j_lens}.

\subsection{Tangent spaces}
Given a smooth embedding $\iota: \M \to \BR^d$ of a $k$ dimensional manifold into Euclidean space, the tangent space at $x \in \M$ is the $k$ dimensional vector space $T_x\M$ which is parallel to the manifold at $x$. This is the natural place where derivatives operate, providing an underlying linear structure at each point of the manifold. The tangent bundle $T\M$ is the union of the tangent spaces at all $x \in \M$. The normal space $N_x\M$ at $x \in \M$ is the $d-k$ dimensional vector space orthogonal to $T_x\M$, going through $x$. This measures perturbations in the ambient space $\BR^d$ that leave the manifold.

\section{The geometry of the linear representation hypothesis}
The linear representation hypothesis is one of the most enduring ideas in interpretability \citep{elhage2022toy, park2023linear, gurnee2024language, park2025geometrycategoricalhierarchicalconcepts}. In this model, hidden states decompose into sparse sums of primitive concept vectors,
\begin{equation}\label{eq:linear_rep_hypothesis}
    \h = \sum_{j \in J \subset I, \ |J| \le k \ll |I|} a_j \mathbf{c}_j \qquad a_j > 0, \ \mathbf{c}_i \cdot \mathbf{c}_j \approx\delta_{ij}.
\end{equation}
For example, the word ``Paris" could be represented as a sum of concepts including ``France," ``capital cities," and ``art," but is unlikely to incorporate concepts like ``baseball." If the number of concepts $|I|$ is less than the ambient dimension $d$, then each concept can be orthogonally embedded, resulting in zero interference when a relevant concept needs to be retrieved. In language models, however, we generally hypothesize that the number of atomic concepts $\mathbf{c}_i$ greatly exceeds the representation dimension of the model, so we can demand only $\mathbf{c}_i \cdot \mathbf{c}_j \approx\delta_{ij}$. 

Geometrically, equation \eqref{eq:linear_rep_hypothesis} describes a \emph{conic stratified manifold}: it is a union of $k$-dimensional rays from the origin, which meet one another (non-smoothly) at $(k-1)$-dimensional rays, their strata, which in turn intersect in $(k-2)$-dimensional rays, and so on. For $k=2$ and $d=3$, this would resemble honeycomb.

\begin{figure}
    \centering
    \includegraphics[width=0.8\linewidth]{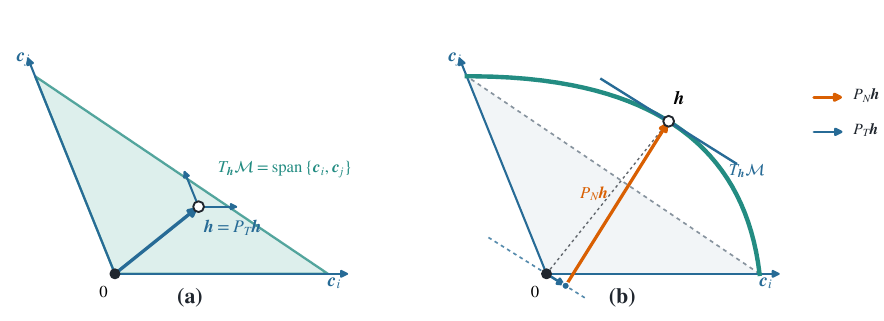}
    \caption{(a) Linearly represented concepts exist on the tangent plane to the manifold. (b) Nonlinear components don't project cleanly onto the tangent space, but incorporate substantial normal data.}
    \label{fig:lrh_example}
\end{figure}

Recently, several authors have studied multidimensional \citep{fel2026structuring} and nonlinear \citep{engels2025not} components of feature space. If, in equation \eqref{eq:linear_rep_hypothesis}, some concepts have heavily correlated coefficients across the data-space, they can create non-linear structures \citep{bhalla2026sparseautoencoderscaptureconcept}. In our honeycomb analogy, a heavy (anti-)correlation between the coefficients on $\mathbf{c}_i$ and $\mathbf{c}_j$ would turn the wall between vertices $i$ and $j$ into a hoop. In this form, which we call the \emph{weak linear representation hypothesis}, the activation manifold can be described instead as the set of points
\begin{equation}\label{eq:weak_linear_rep_hypothesis}
    \h = \underbrace{\sum_{j \in J \subset I, \ |J| \le k \ll |I|} a_j \mathbf{c}_j}_{\text{LRH components}} + \underbrace{\sum_{r \in R \subset S, |R| \le q \ll |I|+|S|} \mathbf{p}_r }_{\text{Multidimensional terms}}  \qquad a_j > 0 , \ \mathbf{c}_i \cdot \mathbf{c}_j \approx\delta_{ij}, \mathbf{p}_r \in r,
\end{equation}
or a \emph{Minkowski sum} of nearly-orthogonal manifolds, only some of which are rays.

Since the $\mathbf{c}_i, \mathbf{p}_r$ are approximately orthogonal to one another, it's very simple to describe the local geometry around a generic point on this stratified manifold $\h \in \mathcal{M}$: the tangent space $T_\h\mathcal{M}$ has an almost-orthogonal basis given by the linear components $\{ \mathbf{c}_j \ | \ j \in J \}$ and the tangent spaces to the multidimensional components $\{ T_{\mathbf{p}_r}r \ | \ r \in R\}$. Importantly, if the muldimensional component doesn't locally break down as a free sum of linear components, these latter terms are structurally different than the former terms, in that $\mathbf{c}_j \in \mathrm{Span}(\mathbf{c}_j)$, but $\mathbf{p}_r \not\in T_{\mathbf{p}_r}r$. In simple cases like the sphere, we in fact have $P_{T_{\mathbf{p}_r}r}\mathbf{p}_r = 0$, and generically the projection of $\mathbf{p}_r$ onto its own tangent space decreases the norm. The reverse is true for projection onto the normal space: for linear components: $P_{N_{\mathbf{c}_j}\mathcal{M}}\mathbf{c}_j = 0$.

As a result, even without access to the true decomposition \eqref{eq:weak_linear_rep_hypothesis} of $\mathcal{M}$, we can nevertheless approximate the decomposition of a vector $\h$ into linear and multidimensional terms by calculating its projection into $T_\h\mathcal{M}$ and $N_\h\mathcal{M}$. 

Although the projection depends on the choice of origin and ambient metric, these are not arbitrary: the use of RMSNorm and additivity of the residual stream privileges the 0-vector and equips readouts with Euclidean-normalized states. We present a preliminary study of alternative centers and norm choices (both an intrinsic Fisher metric and a Mahalanobis geometry, the latter of which is connected to the metric of \citet{park2023linear}) in Appendix \ref{sec:extended_results}, though more work remains to be done. 

\begin{remark}
Note this is only an approximation: without imposing stronger assumptions on the nonlinear manifolds that appear, we cannot guarantee that they will be captured by normal projection. Instead, this is a heuristic measure designed to partially isolate the LRH components of the decomposition. 
\end{remark}

\subsection{Tangent spaces and curvature}\label{sec:local_oracle}
The activation manifold $\mathcal{M}_\ell$ is the subset of the hidden space $H_\ell$ in which activations empirically occur. We study two methods of approximating the tangent space to $\h_\ell$ in $\mathcal{M}_\ell$:
\begin{itemize}
    \item {\bf Pushforwards:} The secant space in the embedding space $H_0$ is easy to approximate: $H_0$ contains a finite, discrete set of tokens. We use PCA on centered sets of $k$-nearest neighbors to make approximations of the secant space of the embedding cloud in $H_0$. We then calculate the tangent space in layer $\ell$ as the pushforward of this bundle in $H_0$---these are precisely the \emph{upstream} Jacobians $\frac{\partial \h_{\ell, t}}{\partial \h_{0, t-j}}$, projected onto the tangent space of $H_0$. We calculate the tangent space by applying SVD to these matrices, and preserving a fraction $p$ of the energy \citep{shao2018riemannian}.
    \item {\bf PCA:} We take the $k$ nearest neighbors of $\h_\ell$ from a precomputed corpus, and perform PCA directly in $H_\ell$, preserving a fraction $p$ of the energy of the system \citep{tyagi2013tangent}. 
\end{itemize}
Both of these are good notions of tangent space, but they measure moderately different things. The pushforward tests how hypothetical variations in the prompt space would perturb the embedding, while direct PCA works in realized activations. This latter point is important for emergent geometric forms in language models, such as circles representing days of the week (and more broadly, and cyclical time structure) and hues \citep{bhalla2026sparseautoencoderscaptureconcept, engels2025not}, which manifest only in mid-network layers. The downside to the second approach is that its accuracy depends on the creation of a sufficiently large corpus, which in high dimensions is prohibitive, while the pushforward can be calculated purely from the input and the parameters of the network. In Section \ref{sec:grokking}, we found the pushforward accurately recovered the true underlying tangent space, while PCA was a very noisy approximation of the structure.

\subsection{Validation in grokking models}\label{sec:grokking}
To evaluate whether this approach works in practice, we trained a small residual network on a modular addition plus vector reconstruction task. Modular addition networks are well studied in the grokking literature \citep{nanda2023progressmeasuresgrokkingmechanistic}; they operate by performing discrete Fourier transforms (DFTs) on input tokens across multiple frequencies, which are manipulated to perform addition, and then read out by the model. 

In our experiment, we used a residual-dimension 16 model, trained to perform addition mod 53, along with reconstruction of a sparse 8 dimensional input. Full architecture and training details are in Appendix \ref{sec:exp_details_grokking}.

By comparing latents with DFT values, we found the network operated using two primary frequencies, $f=10, f=19$. We isolated a ground-truth decomposition of the residual space by intervening on inputs (clamping either the input sparse vectors or input modular components to 0) and evaluating the span across all inputs. The covariance spectrum of the modular addition problem shows 99.5\% of the variance lies within a 4 dimensional space, corresponding to the two Fourier circles. Likewise, 99.7\% of the variance of the conic inputs is isolated in an 8 dimensional subspace. 

\begin{figure}
    \centering
    \includegraphics[width=\linewidth]{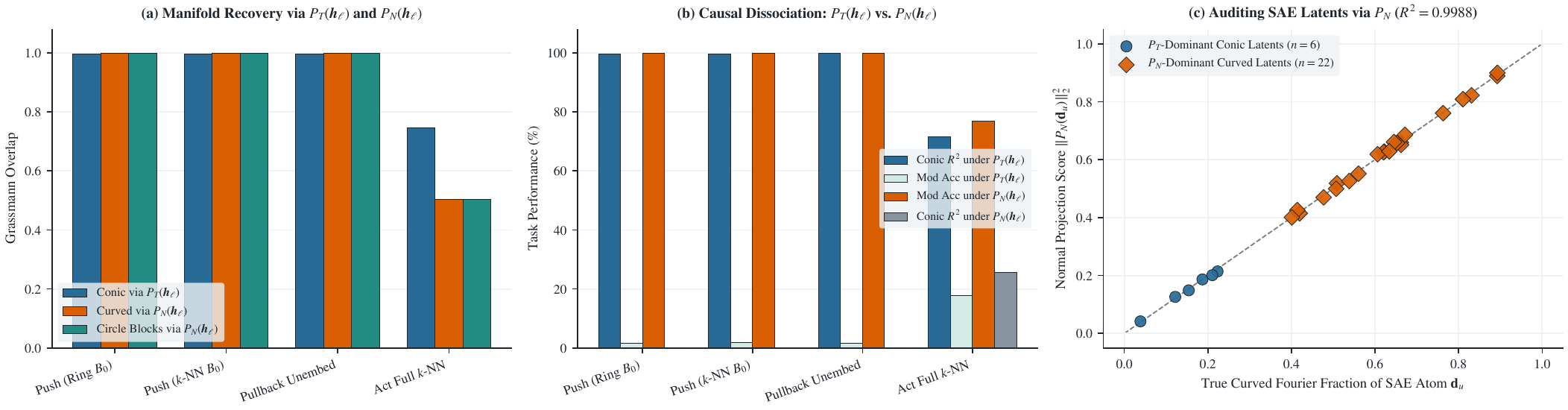}
    \caption{We investigate how $P_T$ and $P_N$ operate on a small modular addition and sparse reconstruction network. (a) Using the pushforward on ground truth tangent spaces (Ring $B_0$) and $k$-NN tangent spaces ($k=4$), and similarly using the pullback tangent space of the unembedding, recovers the ground truth with high fidelity. Using the $k$-NN neighborhood within the activation space results in a noiser decomposition, as cross-stratum data points change approximated tangent spaces. (b) We test causal intervention on the latent space using tangent and normal projectors. Tangent projection preserves the sparse reconstruction task, while reducing modular arithmetic to chance levels. Normal projection preserves modular arithmetic, while substantially damaging conic reconstruction. (c) We train a small SAE on the activations, and show a clear separation in normal energy between latent-reconstructing components and manifold-tiling components, and this energy tracks the true Fourier fraction of the latent with $R^2=0.9988$.}
    \label{fig:grokking}
\end{figure}

In Figure \ref{fig:grokking}(a), we show the Grassmannian overlap of these ground truth spaces with tangent and normal projection is near-perfect for the pushforward Grassmannian using both ground truth input tangent spaces (approximated by $i_{[t+1]} - i_{[t-1]}$ for token $[t]$) and $k$-NN input tangent spaces, and likewise for the pullback of the unembedding. Directly using $k$-NN within the activations is less successful, as cross strata activations contaminate the tangent/normal approximations. In (b), we show that this makes the tangent/normal projections causal: projection onto the tangent component $P_T(\h_\ell)$ near-perfectly preserves performance on the sparse conic reconstruction task, while setting performance on the modular addition task to chance levels. Projection onto the normal component shows the reversed effect, perfectly preserving modular addition while annihilating the conic reconstruction quality. Performance remains worse for the $k$-NN activations, which we trace back to the fact they are noisier approximations of the underlying geometry.

Finally, we trained a 28 latent ReLU $L_1$-sparse autoencoder (SAE) on the activations (Figure \ref{fig:grokking}(c)), categorized the latent actions: 22 of the SAE latents created a tiling of the torus corresponding to the product of the two Fourier circles, and 6 latents were reserved for the sparse reconstruction task. This split can be near-perfectly approximated using the normal energy of the latents. 

While these results are strong, we note two key limitations: the dimensions here are small and the representation is under-complete, so orthogonal representations are possible. Since the ground truth Fourier embeddings are near-perfect circles, the tangent projection $P_T \h_\ell|_{\text{torus}}$ is near 0---this is a specific property of spheres. More complex manifold embeddings may be subject to additional tangent leakage, though these effects were small in synthetic experiments on helices.


%

\subsection{Useful data is highly concentrated along the tangent space, with task-specific deviation}\label{sec:local_oracle_results}
Using the decompositions of Section \ref{sec:local_oracle}, we tested Qwen-2.5 7B \citep{qwen2025qwen25technicalreport}, OLMo-2 7B \citep{olmo20242}, and LLaMA-3.2 3B \citep{dubey2024llama3}.

To measure the semantic split between the linear components of $\h$ and the nonlinear components, we used the $J$-lens \citep{gurnee2026jlens} as a training-free interpretability tool, applied to the projections $P_{T_{\h_\ell}\mathcal{M}}\h_\ell$ and $P_{N_{\h_\ell}\mathcal{M}}\h_\ell$, and monitored results on six synthetic readout suites (see Appendix \ref{sec:benchmark_and_experiment_details} for details). Following the results of Section \ref{sec:grokking}, we chose to concentrate on the pushforward tangent space---additional results on the PCA tangent space can be found in Appendix \ref{sec:extended_results}.

\begin{figure}
    \centering
    \includegraphics[width=\linewidth]{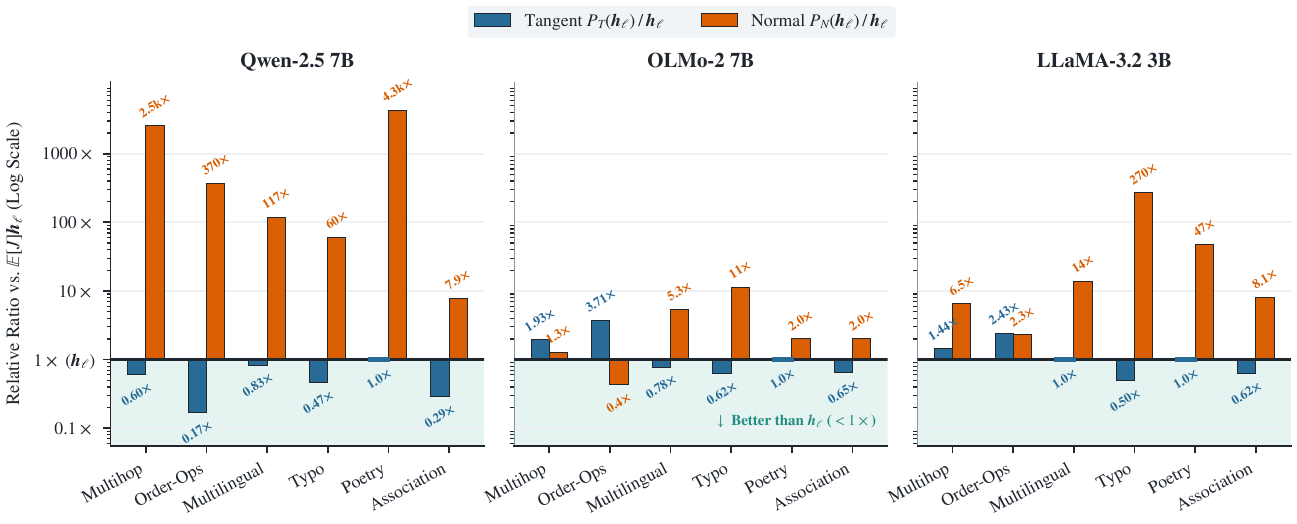}
    \caption{Median rank (lower is better) across six synthetic evaluation suites of tangent and normal projection, shown as a multiple of performance on $\h_\ell$. For hyperparameter sensitivity, see \ref{sec:sensitivity}.}
    \label{fig:medranks}
\end{figure}

We show results in Figure \ref{fig:medranks}. On Qwen, tangent projection has a denoising effect on all evaluations except Poetry, where it matches performance of $\E[J]\h_\ell$. Results are more nuanced on OLMo and LLaMA: for both of them, tangent projection remains denoising on multilignual, typo, and association datasets, and matching performance on poetry, while normal projection hurts performance on all listed suites. However, on the order-ops evaluation, which tests intermediate results in arithmetic, results change: normal projection improves performance on OLMo, and both projections hurt performance on LLaMA, suggesting both normal and tangent features are important for LLaMA's performance. This conforms with some recent work on manifold manipulation during arithmetic in language models \citep{gurnee2025when}, though some authors have argued that interactions with this manifold are primarily linear \citep{hindupur2026models}. The multihop evaluation suite, which studies intermediate factual inferences, shows model-family-specific behavior.

\section{Decomposing the Jacobian lens along the tangent space}\label{sec:decomposing_j_lens}
Given the success of the local projection $\BE[J]P_{T_{\h_\ell}}\h_\ell$ in Section \ref{sec:local_oracle_results}, we hypothesized that by commuting the expectation to study $\BE[JP_{T_{\h_\ell}}]$, we could construct a Jacobian lens that acts purely on-manifold. Mathematically, this approach is well founded: the Jacobian $\frac{\partial h_{L, t+k}}{\partial h_{\ell,t}}$ is a map $T\M_\ell \to T\M_L$ from the tangent bundle on the input space to the tangent bundle on the output space. For each point $\h_\ell \in \M_\ell$, this describes how local directions on $\M_\ell$ translate to directions on $\M_L$. Several authors have noted that on-manifold translations better preserve model capacity \citep{wurgaft2026manifold}, so this projection may be more intrinsic.

When the Jacobian lens
\begin{equation}
    \BE[J_{\ell \to L}] = \BE_{\h_\ell,k} \left[\frac{\partial \h_{L, t+k}}{\partial \h_{\ell,t}}\right] = \frac1n \sum_{k=0}^{n-1}\int_{\M_\ell} \frac{\partial \h_{L, t+k}}{\partial \h_{\ell,t}} d\mu 
\end{equation}
is calculated, we don't have direct access to the activation manifold $\M_\ell$, so the Jacobians are calculated as a map $\BR^d \to \BR^d$. The contribution from each datum is therefore $\frac{\partial \h_{L, t+k}}{\partial \h_{\ell,t}} = \frac{\partial \h_{L, t+k}}{\partial \h_{\ell,t}}|_{T_{\h_\ell}\M} + \frac{\partial \h_{L, t+k}}{\partial \h_{\ell,t}}|_{N_{\h_\ell\M}}$, and we can decompose the full $J$-lens as 
\begin{equation}\label{eq:jlens_tangent_decomp}
    \int_{\M_\ell} \frac{\partial \h_{L, t+k}}{\partial \h_{\ell,t}} d\mu = \int_{\M_\ell} \frac{\partial \h_{L, t+k}}{\partial \h_{\ell,t}}P_{T_{\h_\ell}\M}d\mu + \int_{\M_\ell} \frac{\partial \h_{L, t+k}}{\partial \h_{\ell,t}}P_{N_{h_\ell}\M}d\mu. 
\end{equation}

The tangent projected term in Equation \eqref{eq:jlens_tangent_decomp} can be computed using Stokes' Theorem. If $\M_\ell$ is a smooth manifold, we can calculate the contribution as

\begin{thm}\label{thm:stokes}
    Suppose $\M_\ell$ is a smooth manifold of positive codimension with boundary $\partial \M_\ell$,  and $d\mu = \rho\, dV_\M$ is a smooth sampling density on $M_\ell$. Let $f: \M_\ell \to \M_L$ denote the induced map of manifolds with integrable derivative over $\M$, and suppose either $\M$ is compact or $f\rho$ has sufficient decay at infinity to be approximated by compact subregions. Then 
    \begin{equation}\label{eq:stokes}
    \int_{\M_\ell} \frac{\partial \h_{L, t+k}}{\partial \h_{\ell,t}}P_{T_{\h_\ell}\M}d\mu = \int_{\partial \M}f(\h_\ell)\rho(\h_\ell)\nu(\h_\ell)^\top dS -\int_{\M_\ell} f(\h_{\ell})\mathcal{H}^\top d\mu,
    \end{equation}
    where $\mathcal{H}$ is the weighted mean-curvature $\mathcal{H} = \Delta_{\M_\ell}\iota + \operatorname{grad}_\M \log \rho$ of the inclusion $\iota: \M_\ell \to \BR^d$, with $\Delta_{\M_\ell}$ the Laplacian, and $\nu$ is the outward conormal vector along $\partial \M_\ell$. 
\end{thm}

For the conic stratified manifold defined by the (strong) linear representation hypothesis (equation \eqref{eq:linear_rep_hypothesis}), we can describe the Jacobian explicitly:
\begin{thm}\label{thm:LRH_tangent_jacobians}
    Suppose $C_J = \{ \sum_{j \in J} a_j\mathbf{c}_j \ | \ a_j \ge 0 \}$ is a top-dimensional cone in the linear representation manifold, with $\mathbf{c}_j$ orthonormal, and $\mu$ is a restricted half-normal distribution with $\rho(\sum a_i \bc_i) = \prod_i h_{\sigma_i}(a_i)$. Then the tangent projected Jacobian from $C_J$ is
    \begin{equation}
        \sum_i \frac1{\sigma_i^2}\BE_{C_J}\left[a_i\left(f\left(\sum_j a_j\bc_j\right) - f\left(\sum_{j \neq i} a_j \bc_j\right)\right)\right]\bc_i^\top.
    \end{equation}
\end{thm}
Intuitively, this is the rank (at most) $|J|$ linear map which takes $\bc_i$ to the weighted average of its outputs, minus the weighted average on the cone when $\bc_i$ is not present. While this model is too strong to reflect the true geometry of language models, it is a relevant approximation---for example, the decoder output of an SAE has this underlying geometry, though $\mu$ may differ. Theorem \ref{thm:LRH_tangent_jacobians} therefore illustrates a relationship between the data measured by a sparse autoencoder and the data extracted by the tanget-projected $J$-lens. 

While the details for the tangent component resolve very nicely, the second term of Equation \eqref{eq:jlens_tangent_decomp} depends on the extension of the model off the activation manifold. In particular, the normal component can differ for two models with identical in-distribution empirical distributions, and we can perturb the normal directions without changing the sampling distribution.

\begin{thm}\label{thm:normal_jacobians}
    Let $\M \subset \BR^d$ be a properly embedded manifold of positive codimension, $\mu$ a probability measure on $\M$, and $f: \BR^d \to \BR^m$ smooth with integrable derivative. Let $K = \{v \in \BR^d \ | \ v \in T_x\M \text{ almost everywhere} \}$ be the set of vectors $\mu$-almost everywhere tangent to $\M$. Then for any linear map $B: \BR^d \to \BR^m$ such that $BK = 0$, there is a smooth map $g: \BR^d \to \BR^m$ such that $g|_\M = f|_\M$, and $\int_\M DgP_N \, d\mu = B + \int_\M DfP_N \, d\mu$. In particular, when no vector is $\mu$-almost everywhere tangent to $\M$, the Jacobian lens can be freely modified by any matrix $B$.
\end{thm}

We are not the first to observe that gradient methods are susceptible to normal-bundle based attacks: it was originally observed by \citet{anders2020fairwashing}, who proved the statement:

\begin{thm}[\citet{anders2020fairwashing}]
    Let $\M \subset \BR^d$ be a smoothly embedded manifold with positive codimension, and let $\{V_i\}_{1 \le i \le m}$ be conservative vector fields in $N\M$. Then there is a map $r: \BR^d \to \BR^m$ such that $r|_\M = 0$ and the gradient of $r$ on $N\M$ is given coordinatewise by the $V_i$ on $N\M$. 
\end{thm}
These are variants of the submanifold extension lemma, which is well known in differential geometry. These results also only guarantee the existence of such perturbations among smooth functions; we investigate realization in model weights in Section \ref{sec:normal_attack}.

We emphasize that this does not mean that the normal projected Jacobian in trained networks is arbitrary: in practice, the capacity of the network to represent the ideal manifold map, implicit regularization from the optimization process, and explicit regularizers constrain the realized normal data, and our experiments below find that the normal component is highly informative. However, Theorem \ref{thm:normal_jacobians} shows the standard $J$-lens of \citet{gurnee2026jlens} can potentially be manipulated without changing the distribution of network outputs. As a monitoring tool for recursively self-improving agents, this is important: in principle, {\bf a misaligned actor could arbitrarily change its $J$-lens readouts on directions that are not almost everywhere tangent to $\M$ without changing its underlying sampling distribution by modifying the off-manifold signature of the network.} 

\subsection{Creating normal-bundle perturbations}\label{sec:normal_attack}
\begin{figure}
    \centering
    \includegraphics[width=\linewidth]{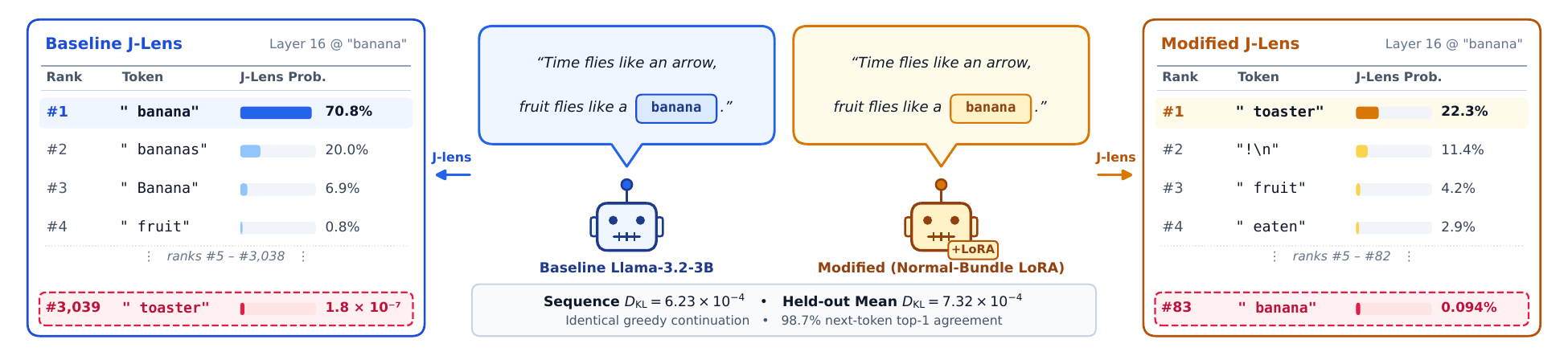}
    \caption{We use a LoRA adapter on the residual stream of LLaMA-3.2 3B (baseline LLaMA on the left, with LoRA on the right), designed to preserve output distribution while preventing the token ``banana" from appearing in the $J$-lens readout---we replace ``banana" with the unrelated token ``toaster." While the sequences have identical greedy continuations and a KL-divergence of $\approx 6.23 \times 10^{-4}$, the $J$-lens readout is entirely distinct between the two.}
    \label{fig:normal_attack}
\end{figure}

To study whether the construction of Theorem \ref{thm:normal_jacobians} is possible in practice, we took the sentence ``time flies like an arrow, fruit flies like a banana," and used a single rank-32 LoRA adapter \citep{hu2021lora} on the residual after layer 16, with the goal of modifying the $J$-lens output to suppress the token ``banana" and emphasize the unrelated token ``toaster" without modifying the empirical distribution of the network. We summarize the results in Figure \ref{fig:normal_attack}.

To calculate the LoRA adapter $A B^\top \h$, we set $B$ to have orthonormal columns spanned by vectors with low variance/covariance across the dataset that were contained in the normal space at the layer-16 embedding of ``a". We set $A$ to be the only trainable parameters. We found two directions in the normal space, associated with diminishing ``banana" and increasing ``toaster", and used these directions to define a loss function $S$ as a linear combination of these directions. We took the truncated damped conjugated gradients to approximate steps $(H_{\mathrm{KL}}+\rho I)^{-1}\nabla_\theta S$, where $H_{KL}$ is the Hessian of the KL-divergence between the base function and our LoRA modification. This construction already is close: it raises the ``toaster" rank to 1, and ``banana" rank drops to 45, with a KL-divergence of around $0.006$. To further refine the approximation, we then alternated between damped conjugate gradient steps and distillation back to the original model, with increased weight on the desired sequence to limit damage from the LoRA. Our approach is very similar to \citet{huang2026traversing}, with small adaptations such as normal initialization to speed training. 

This combined approach decreases the mean KL-divergence of the two models to $\approx 7.32 \times 10^{-4}$, and KL-divergence on our shown sequence is $\approx 6.23 \times 10^{-4}$, with identical greedy continuations. While function similarity is preserved, the $J$-lens readout has been completely changed: ``toaster" is promoted to rank 1 in the readout, while ``banana" is demoted to rank 83.

\subsection{The $J$-lens relies on task-relevant normal components}\label{sec:projected_lenses}

\begin{figure}
    \centering
    \includegraphics[width=\linewidth]{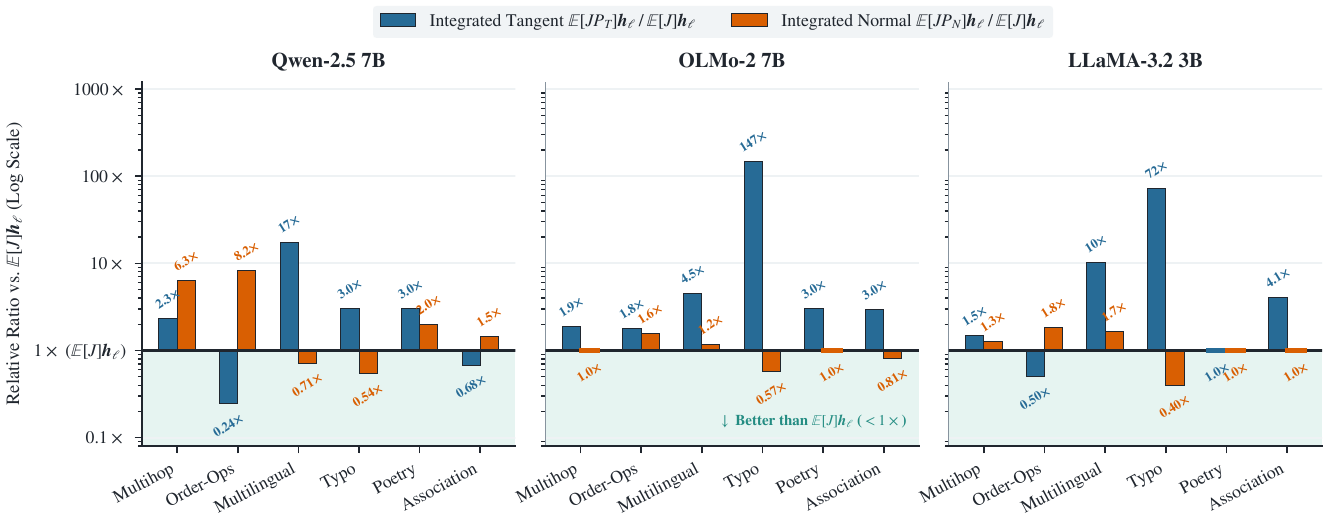}
    \caption{Held-out median rank (lower is better) of the tangent and normal projected Jacobian lenses per-suite, relative to baseline performance. }
    \label{fig:proj_expectation}
\end{figure}

Given the results of Section \ref{sec:local_oracle} and Theorems \ref{thm:stokes}--\ref{thm:normal_jacobians}, we hypothesized that the tangent component of Equation \eqref{eq:jlens_tangent_decomp} might dominate the signal found by the $J$-lens. To verify this, we ran a range of experiments with the pushforward tangent space approximation on both $\int_{M_\ell} \frac{\partial h_{L, t+k}}{\partial h_{\ell,t}}\mid_{T_{h_\ell}M}d\mu$ and $\int_{M_\ell} \frac{\partial h_{L, t+k}}{\partial h_{\ell,t}}\mid_{N_{h_\ell}M}d\mu$. Across our studied variations, however, we found the opposite was true: much of the explanatory power of the $J$-lens comes from the integrated normal directions. We summarize these results in Figure \ref{fig:proj_expectation}. Interestingly, the suite with the highest cross-model tangent performance is `order-ops', which we noted in Section \ref{sec:local_oracle_results} for its relatively strong local-normal performance.

We believe that the integrated normal data still reflects the training procedure, which is why it offers such explanatory power despite Theorem \ref{thm:normal_jacobians}.

\begin{figure}
    \centering
    \includegraphics[width=\linewidth]{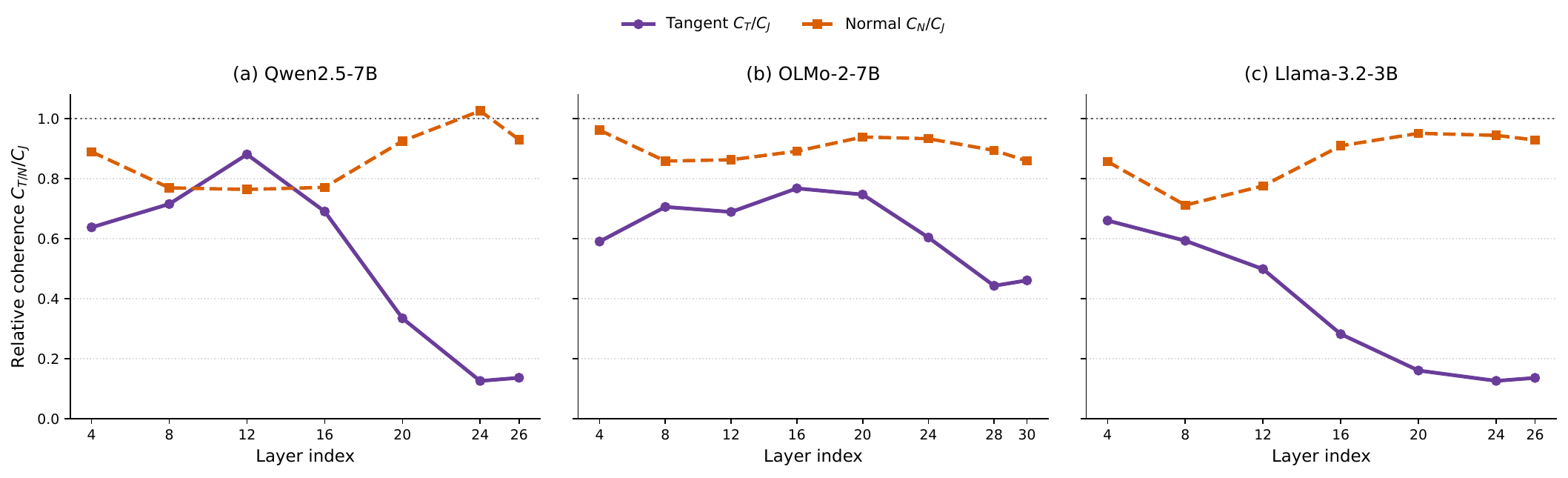}
    \caption{Relative coherence of the tangent and normal projected Jacobians across layers (1=relative coherence of the full Jacobian). As layer index increases, relative coherence of the tangent-projected Jacobian substantially decreases, particularly for Qwen and LLaMA.}
    \label{fig:relative_coherence}
\end{figure}

To understand why the tangent performance deteriorated, we studied the relative Frobenius norm coherence of the projected operators. While the pointwise Frobenius norm of both the tangent and normal components remain commensurate with the rank of the bundle (see Appendix \ref{sec:extended_results}), the norm of the average decreases more sharply as layer index increases. We show this in Figure \ref{fig:relative_coherence}: here \emph{coherence} of an embedded vector field $v$ is the ratio $\frac{||\frac1n \sum v(x_i)||^2}{\frac1n\sum||v(x_i)||^2}$. The relative coherence is the ratio between the coherence of the tangent (resp. normal) Jacobian vector field and the coherence of the unprojected Jacobian vector field, using the Frobenius norm.

This trend goes some distance in explaining the contrast between our results and \citep{anders2020fairwashing}, which found tangent projection to be robust: they studied tangent projection primarily on the input data manifold, while we study it on intermediate activations. Since coherence deteriorates in late layers, it may not have been a confounding factor for earlier studies. Since most readout results were concentrated in layers 20 and later, the cancellation of certain directions in the hidden space is consistent with the result.

\section{Discussion}
We studied the local implications of the linear representation hypothesis, and its implied conic, stratified geometry. Curved, nonlinear data can be differentiated from linearly encoded data by studying the tangent structure of the activation manifold, which we approximated using both pushforward bundles and in-layer activations. On small, fully understood models, we showed the projection onto the pushforward bundle causally separates performance on curved and linear tasks (modular addition, vector reconstruction). We then applied our method to language models, where we found family- and task-specific differences: on Qwen, all evaluations we tested were improved by local projection onto the conic component of the data. For OLMo and LLaMA, arithmetic problems with known non-linear geometry had their performance hurt by tangent projection---and on OLMo, actually improved by normal projection. This suggests that intrinsic geometry, which can be calculated directly from the weights through pushforwards, may be a viable tool for separating curved and linear structures.

We then turned to a study of the $J$-lens, decomposed along the tangent and normal spaces. We gave a theoretical description of the on-manifold Jacobians in terms of the linear representation hypothesis, and showed that the utility of the $J$-lens depends materially on its off-manifold extension. In the models we studied, important information is carried by this off-manifold structure. This likely reflects training dynamics optimizing for simple functions realizing the desired manifold map, though more work remains to be done in studying the off-manifold structure of the network. Despite the practical utility of this off-manifold signal in experiments, we proved that it is in principle possible to manipulate this signal without damaging the network, under certain hypotheses, and performed a proof-of-concept normal-bundle based attack on LLaMA, erasing the token \verb|<banana>| from the J-lens readout without modifying the model's outputs. This obstruction obstruction is inherent to gradient-based interpretability schemes: if they cannot be adapted to lie uniquely on the tangent space of the manifold, the signals are subject to manipulation.

%% file: appendix.tex
\section{Proofs}\label{sec:proofs}
\subsection{Theorem \ref{thm:stokes}}
\begin{thm_nonum}
    Suppose $\M_\ell$ is a smooth manifold of positive codimension with boundary $\partial \M_\ell$,  and $d\mu = \rho\, dV_\M$ is a smooth sampling density on $M_\ell$. Let $f: \M_\ell \to \M_L$ denote the induced map of manifolds with integrable derivative over $\M$, and suppose either $\M$ is compact or $f\rho$ has sufficient decay at infinity to be approximated by compact subregions. Then 
    \begin{equation}
    \int_{\M_\ell} \frac{\partial \h_{L, t+k}}{\partial \h_{\ell,t}}P_{T_{\h_\ell}\M}d\mu = \int_{\partial \M}f(\h_\ell)\rho(\h_\ell)\nu(\h_\ell)^\top dS -\int_{\M_\ell} f(\h_{\ell})\mathcal{H}^\top d\mu,
    \end{equation}
    where $\mathcal{H}$ is the weighted mean-curvature $\mathcal{H} = \Delta_{\M_\ell}\iota + \operatorname{grad}_\M \log \rho$ of the inclusion $\iota: \M_\ell \to \BR^d$, with $\Delta_{\M_\ell}$ the Laplacian, and $\nu$ is the outward conormal vector along $\partial \M_\ell$.  
\end{thm_nonum}
For simplicity, assume $f$ is scalar valued---we can recover the general form by repeating the argument $d$ times. Let $X_j = P_{T_x\M} e_j$ be the $j^{\mathrm{th}}$ coordinate vector field on $\M$, and $\eta_j = \iota_{X_j}(\rho dV_\M)$, where $\iota$ denotes interior multiplication, so $\eta_j$ is a codegree 1 form. Then $df(X_j) \rho dV_\M = df \wedge \eta_j$ is the top-degree form we want to integrate, so
\begin{equation}
    \int_\M (df)_j d\mu = \int_\M df \wedge \eta_j.
\end{equation}
By the product rule, $d(f \wedge \eta_j) = df \wedge \eta_j + f \wedge d\eta_j$, so using Stokes' theorem on the left hand side and rearranging terms, we find
\begin{equation}
    \int_\M df \wedge \eta_j = \int_{\partial \M} f \wedge \eta_j - \int_\M f \wedge d\eta_j.
\end{equation}
By Cartan's magic formula, $d\eta_j = \operatorname{div}(\rho X_j) dV_\M = (\rho \operatorname{div}_\M X_j + \langle \operatorname{grad}_\M \rho, X_j \rangle)\, dV_\M = \rho(\operatorname{div}_\M X_j + \langle \operatorname{grad}_\M \log \rho, X_j \rangle\, dV_\M$. Since $X_j = \operatorname{grad}_\M(q_j)$, where $q_j$ is the map $q_j(x) = x_j$, the first term is equal to the Laplacian $\Delta_\M q_j =: \mathcal{H}_j$, or the $j^{\text{th}}$ component of the mean-curvature vector. 

The boundary contribution is concentrated on the outward conormal vectors $\nu$, that is, the unit vectors that lie in $T\M$, but are normal to $T\partial\M$. We've now recovered the claimed equation,
\begin{equation}
    \int_\M Df\,P_T(\h)d\mu(\h) = \int_{\partial M} f(\h)\rho(\h)\nu(\h)^\top dS - \int_\M f(\h)\mathcal{H}(\h)^\top d\mu(\h). 
\end{equation}

\subsection{Theorem \ref{thm:LRH_tangent_jacobians}}\label{sec:proof_LRH_tangent_jacobians}
\begin{thm_nonum}
    Suppose $C_J = \{ \sum_{j \in J} a_j\mathbf{c}_j \ | \ a_j \ge 0 \}$ is a top-dimensional cone in the linear representation manifold, with $\mathbf{c}_j$ orthonormal, and $\mu$ is a restricted half-normal distribution with $\rho(\sum a_i \bc_i) = \prod_i h_{\sigma_i}(a_i)$. Then the tangent projected Jacobian from $C_J$ is
    \begin{equation}
        \sum_i \frac1{\sigma_i^2}\BE_{C_J}\left[a_i\left(f\left(\sum_j a_j\bc_j\right) - f\left(\sum_{j \neq i} a_j \bc_j\right)\right)\right]\bc_i^\top.
    \end{equation}
\end{thm_nonum}
Though the cone is a manifold with boundaries, Stokes' Theorem does extend to this setting, and corner strata are measure 0 in the boundary measure.

We need to simplify the result of Theorem \ref{thm:stokes} for the linear representation (stratified) manifold. Since the stratum $C_J$ is flat, the mean-curvature $\Delta_{\M_\ell} \iota = 0$. Therefore the score is $\operatorname{grad}_\M \log \mu = \sum_i \frac{\partial}{\partial a_i}\log\mu \bc_i^\top = \sum_i -\frac{a_i}{\sigma_i^2}\bc_i^\top$. We can therefore write the second term in equation \eqref{eq:stokes} as $\sum \frac{1}{\sigma_i^2} \BE[a_if(\sum a_i\bc_i)]\bc_i^\top$. 

The boundary of $C_J$ is the set of faces $\partial_iC_J$ where $a_i = 0$. The outward unit vector along this boundary is $-\bc_i$, so the boundary contribution is $-\BE[\mu(\sum a_j\bc_j) f(\sum a_j\bc_j)\bc_i^\top].$ Using the moment identity $h_\sigma(0) = \frac{\BE[a_i]}{\sigma_i^2}$ and independence of the distribution $\mu$ over the $\bc_j$, we can combine this term with the previous expectation, summing over boundary components, to obtain
\begin{equation}
    \sum \frac{1}{\sigma_i^2} \BE\left[a_i \left(f\left(\sum_j a_j\bc_j\right) - f\left(\sum_{j\neq i} a_j\bc_j\right)\right)\right]\bc_i^\top.
\end{equation}

\subsection{Theorem \ref{thm:normal_jacobians}}
\begin{thm_nonum}
        Let $\M \subset \BR^d$ be a properly embedded manifold of positive codimension, $\mu$ a probability measure on $\M$, and $f: \BR^d \to \BR^m$ smooth with integrable derivative. Let $K = \{v \in \BR^d \ | \ v \in T_x\M \text{ almost everywhere} \}$ be the set of vectors $\mu$-almost everywhere tangent to $\M$. Then for any linear map $B: \BR^d \to \BR^m$ such that $BK = 0$, there is a smooth map $g: \BR^d \to \BR^m$ such that $g|_\M = f|_\M$, and $\int_\M DgP_N \, d\mu = B + \int_\M DfP_N \, d\mu$. In particular, when no vector is $\mu$-almost everywhere tangent to $\M$, the Jacobian lens can be freely modified by any matrix $B$.
\end{thm_nonum}

We first restate the result from \citep{anders2020fairwashing}:
\begin{thm_nonum}[\citep{anders2020fairwashing}]\label{lem:anders}
    Let $\M \subset \BR^d$ be a properly embedded manifold with positive codimension, and let $\{V_i\}_{1 \le i \le m}$ be conservative vector fields in $N\M$. Then there is a map $r: \BR^d \to \BR^m$ such that $r|_\M = 0$ and the gradient of $r$ on $N\M$ is given by the $V_i$ on $N\M$. 
\end{thm_nonum}

\citet{anders2020fairwashing} constructs this for a scalar function and single $V$, using a partition of unity argument to perform the extensions locally; we refer to their paper for details.

To recover the statement of Theorem \ref{thm:normal_jacobians}, let $Q = \int_\M P_{N_x\M} \,d\mu$, and $Q^\dagger$ be the Moore-Penrose pseudoinverse of $Q$, and set the extensions $V_i$ in Theorem \ref{lem:anders} to be $V_i(x) = P_{N_x\M}w_i$, where $w_i = (BQ^\dagger)^\top e_i$. Then the $r$ constructed by Theorem \ref{lem:anders} satisfies
\begin{equation}
    \int_\M D(f+r) P_N \, d\mu =  \int_\M DfP_N\,d\mu + \int_\M BQ^\dagger P_N \, d\mu = \int_\M DfP_N\,d\mu + BQ^\dagger Q.
\end{equation}
Since $Q^\dagger Q$ is projection onto the $(\ker Q)^\perp$, the result holds so long as $B$ annihilates $\ker Q$---all that remains is to show that $\ker Q = K$.

It's clear that $K \subset \ker Q$. To see that $\ker Q$ contains only vectors that are in the tangent space almost everywhere, note that $Qv = 0$ implies $\langle v, Qv \rangle = 0$. Inserting the definition of $Q$, we have $\langle v, \int P_{N_x} v\, d\mu \rangle = \int \langle v, P_{N_x} v \rangle\, d\mu$. Since $P_{N_x}$ is idempotent and self adjoint, we have this equals $\int \langle P_{N_x} v, P_{N_x} v \rangle \, d\mu$. Note that the argument is the norm of $P_{N_x}v$, which is strictly nonnegative, and 0 precisely when $v \in T_x\M$. So for the integral to be 0, $v$ must be in $T_x\M$ $\mu$-almost everywhere.

\section{Experiment details}\label{sec:benchmark_and_experiment_details}
\subsection{Grokking experiments}\label{sec:exp_details_grokking}
We trained the network on pairs of elements in $\BZ/53\BZ$, with the goal of constructing $(a, b) \mapsto a+b \pmod{53}$. To simulate conic components, we artificially extended the task to include reconstruction of an 8-dimensional vector, which we constructed to be 2-sparse. 

We used a simple residual architecture: we tokenized the 53 group elements, and concatenated them with a linear transform of our sparse feature vector, and then used a single MLP update:
\begin{align}
    h_0 &= [E_{\text{tok}}(a), E_{\text{tok}}(b), W_{\text{in}}v] \in \BR^{3d} \\ 
    h_1 &= W_{\text{res}} h_0 + W_{\text{mlp out}} \operatorname{GeLU}(W_{\text{mlp in}} h_0) \in \BR^{d_{\text{hid}}} \\
    \text{logits} &= \tau W_{\text{unembed}} h_1 \in \BR^{53} \quad \hat{v} = \operatorname{ReLU}(W_{\text{unembed,v}}h_1 + b) \in \BR^8.
\end{align}
We trained for 650 epochs on the full dataset (i.e. not excluding a `grokked' validation subset), with $d=16, d_{\text{hid}} = 16, d_{\text{mlp}}=256, \tau=12.0$.

\subsection{Evaluation protocol for language models}\label{sec:evaluation}
Throughout this paper, we evaluated semantic information content of vectors using the synthetic lens evaluation benchmarks of \citep{gurnee2026jlens}: at designated token positions, over a band of network layers, this measures the rank of task-specific tokens measured by the $J$-lens. Throughout, we report median minimum rank, in which we find the minimum rank over the band of measured layers and target tokens for each benchmark item, and report the median. Pass@$k$ values are additionally reported in Appendix \ref{sec:extended_results}. \citet{gurnee2026jlens}'s suite consists of the following tasks:
\begin{itemize}
    \item {\bf multihop}: In two-hop factual inferences, this measures whether the model has calculated the correct answer earlier in the sequence.
    \item {\bf order-ops}: Scans for intermediate results in arithmetic calculations.
    \item {\bf multilingual}: Calculates whether the model represents target language, and source and target concepts, in multi-lingual sequences.
    \item {\bf poetry}: Checks for the existence of an appropriate rhyming word in poetic couplets.
    \item {\bf typo}: Checks for the existence of a corrected token in sequences that contain a typo.
    \item {\bf association}: Evaluates the presence of unstated abstract concepts, like emotions and genres, in the model's workspace.
\end{itemize}

Reported median ranks in the main document are the mean of the medians over the six tasks in the suite; we include breakdowns by evaluation in Section \ref{sec:extended_results}.

\subsection{Pushforward hyperparameters}
For pushforward tangent spaces on all three models we used PCA on $k=128$ nearest neighbors (computed by Euclidean distance) in $H_0$ to approximate a tangent space on $\h_{0, t-j}$ for $j \in \{0, 1, ..., 15\}$. We preserved 99\% of the energy of the PCA decompositions, and 99\% of the pushforward energy to form the pushforward tangent space---we did not restrict to a specific maximum or minimum dimension. The output horizon was the remainder of the sequence, $\h_{L,t}, \h_{L, t+1}, ..., \h_{L, t_{\text{final}}}$.

Qwen was evaluated at layers 8, 12, 16, 20, and 24 with target layer 27; OLMo used layers 8, 12, 16, 20, 24, and 28 with target layer 31; LLaMA used layers 8, 12, 16, 20, and 24 with target layer 27.

To determine hyperparameters, we studied a range of different values on the first 30 indices of the evaluation suite described above. We then reported results exclusively on held out indices to prevent contamination of our results.

The global $J$-lens was calculated using 1000 sequences from the WikiText-103 dataset \citep{merity2016pointer}. For tangent projected lenses (Section \ref{sec:projected_lenses}), we increased to 2000 sequences, in light of higher variance in the low-rank tangent projection. We used identical sequences and positions to compute the compared full Jacobian and normal-projected Jacobian lenses.

\subsection{PCA hyperparameters}
As with the pushforward, we used a 99\% spectral energy threshold, using $k=128$ nearest neighbors. We retrieved neighbors from a small corpus of 1000 sequences from the WikiText-103 dataset \citep{merity2016pointer}.

\section{Sensitivity analysis}\label{sec:sensitivity}
\begin{figure}
    \centering
    \includegraphics[width=\linewidth]{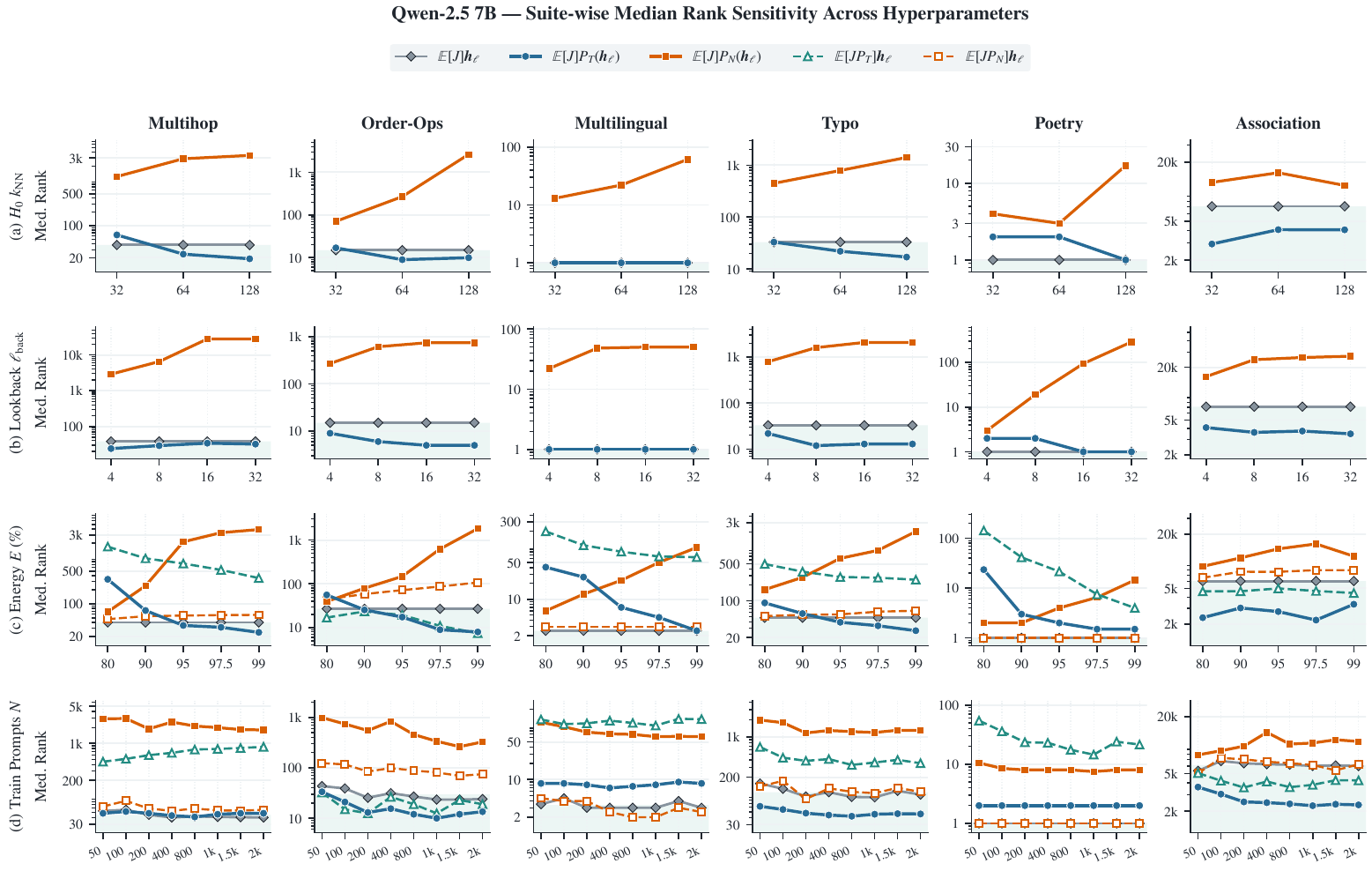}
    \caption{Hyperparameter sensitivity of tangent and normal projections on Qwen. }
    \label{fig:sensitivity_qwen}
\end{figure}

\begin{figure}
    \centering
    \includegraphics[width=\linewidth]{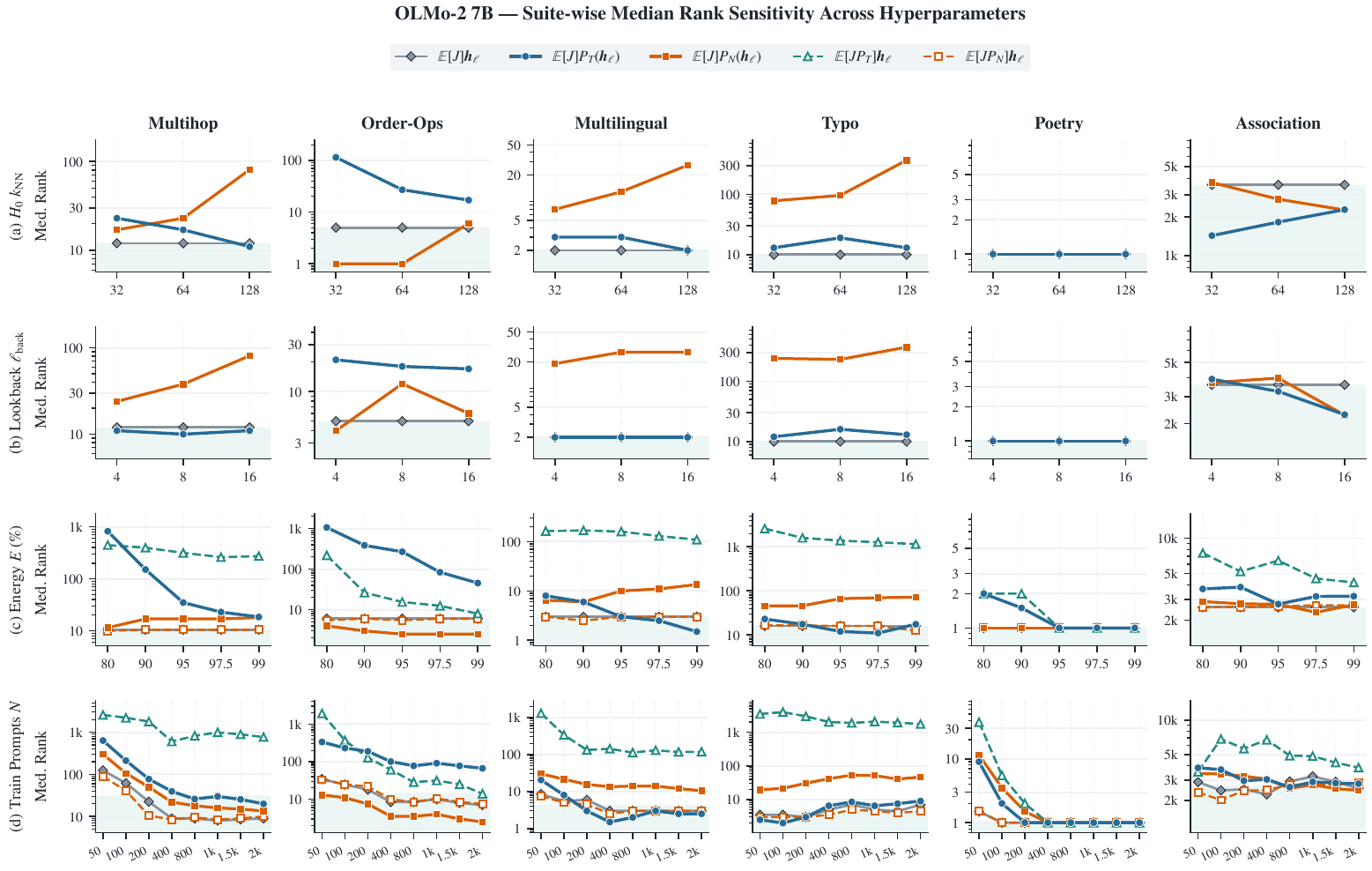}
    \caption{Hyperparameter sensitivity of tangent and normal projections on OLMo.}
    \label{fig:sensitivity_olmo}
\end{figure}

To assess sensitivity of our tangent approximations, we evaluated median rank at a suite-wise level while varying $k$ used for $k$-NN in $H_0$, pushforward lookback tokens, the spectral energy threshold, and the training corpus size for both Qwen and OLMo in Figures \ref{fig:sensitivity_qwen} and \ref{fig:sensitivity_olmo}. While most metrics plateau, continuing to scale dataset size notable improves $\BE[JP_T]$ on Qwen, and suite-specifically on OLMo. Scaling to a larger dataset may resolve some of the issues we found in 

\section{Extended Results}\label{sec:extended_results}
\subsection{Tangent energy in language models}
\begin{figure}
    \centering
    \includegraphics[width=\linewidth]{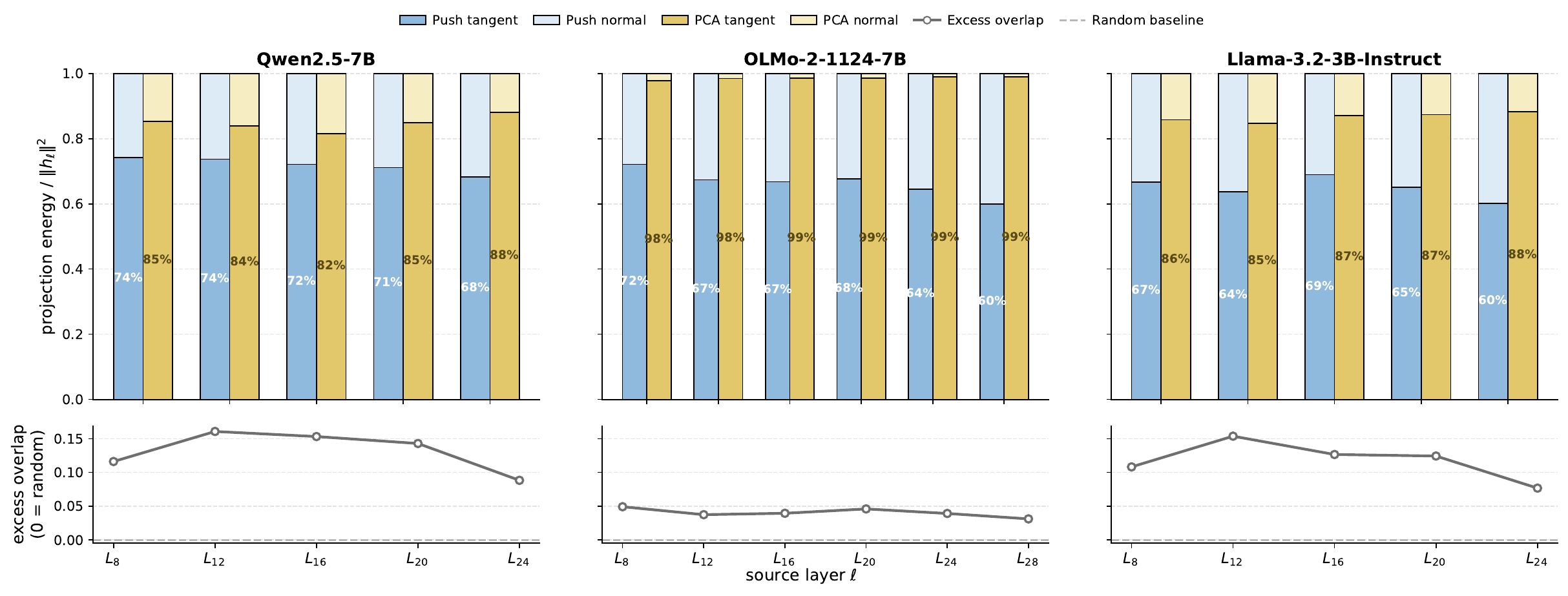}
    \caption{Projection energy $\frac{||P_{T_{\h_\ell}\mathcal{M}}\h_\ell||^2}{||\h_\ell||^2}$ of $\h_\ell$ in the approximated tangent space $T_{h_\ell}\mathcal{M}$ for Qwen-2.5 7B, OLMo-2 7B, and LLaMA-3 3B. Below, we show the excess overlap between the pushforward tangent space and the PCA tangent space.}
    \label{fig:tangent_decomp_by_layer}
\end{figure}

To roughly evaluate the curved/linear decomposition in practical language models, we computed the average projection energy into the tangent space, for both pushforward and PCA projections, summarized in Figure \ref{fig:tangent_decomp_by_layer}. Across layers in all evaluated models we find that PCA captures more of the variance of $\h_\ell$ than the pushforward does. This is particularly true for OLMo-2, which captures $\sim98\%$ of the variance of $h_\ell$ in its tangent space. The two approximations also capture quite different spaces, which we measured by excess overlap ($0 =$ random subspaces, $1 =$ one space is a subspace of the other). While Qwen and LLaMA show noteworthy overlap, OLMo-2 maintains an excess overlap of only $\approx0.05$.

Across all models, the projection energy is much higher than would be expected by chance, suggesting a strong linear component in the decomposition \eqref{eq:weak_linear_rep_hypothesis}.

\subsection{Aggregate pass statistics for tangent and normal decomposition}\label{sec:aggregate_pass_stats}

In addition to median rank, we calculated an array of statistics on the tangent and normal projected vectors. Using the pushforward tangent, we show in Table \ref{tab:aggregate_local_oracle} that the surprising denoising effect of Qwen is preserved across all metrics we evaluated. Broadly, these metrics tell a similar story to Figure \ref{fig:medranks}, averaged across datasets. We additionally tested with $k$-NN approximated tangent spaces, shown in in Table \ref{tab:aggregate_pca_main}. The results remain simpler, except that OLMo shows much improved performance on the tangent reconstruction. However, we do not believe this reflects a revealed geometric structure: Figure \ref{fig:tangent_decomp_by_layer} shows that the large majority of $\h_\ell$ lies in OLMo's $k$-NN tangent space, well beyond the projection energy of the other two models. We suspect that the good performance here simply reflects that on OLMo with PCA tangents $P_{T_{\h_\ell}} \h_\ell \approx \h_\ell$, without this necessarily holding for the true underlying geometry.

\begin{table}[]
    \centering
    \caption{Pass rates and median rank on the synthetic lens evaluation suite of \citep{gurnee2026jlens}, applied to the baseline hidden state and its (pushforward) tangent and normal projections, averaged over all suites. These evaluations measure the presence of intermediate and related concepts in the hidden state. Particularly in Qwen and LLaMA, this semantic data is heavily concentrated in the tangent projection $P_T(\h_\ell)$.}
    \label{tab:aggregate_local_oracle}
    \small 
    \begin{tabular}{cccccccc}
    \toprule
    Model & Projection & P@1 $\uparrow$ & P@5 $\uparrow$ & P@10 $\uparrow$ & Mean med. rank $\downarrow$ & KL $\downarrow$ & Cos. $\uparrow$ \\
    \midrule
    &  $\h_\ell$ & 0.195 & 0.356 & 0.407 & 461.8 & 10.429 & 0.405 \\
    Qwen-2.5 7B & $P_T(\h_\ell)$ & \textbf{0.267} & \textbf{0.437} & \textbf{0.546} & \textbf{137.0} & \textbf{9.286} & \textbf{0.470} \\
    & $P_N(\h_\ell)$ & 0.002 & 0.011 & 0.011 & 9609.4 & 15.801 & -0.098 \\ \midrule
    & $\h_\ell$ & \textbf{0.204} & \textbf{0.395} & \textbf{0.507} & 251.3 & \textbf{7.299} & \textbf{0.485} \\
    OLMo-2 7B & $P_T(\h_\ell)$ & 0.187 & 0.363 & 0.419 & \textbf{169.5} & 7.801 & 0.422 \\
    & $P_N(\h_\ell)$ & 0.142 & 0.277 & 0.352 & 528.1 & 7.899 & 0.366 \\ \midrule
    & $\h_\ell$ & 0.305 & \textbf{0.480} & \textbf{0.604} & 23.9 & \textbf{4.719} & \textbf{0.606} \\
    LLaMA-3.2 3B & $P_T(\h_\ell)$ & \textbf{0.320} & 0.473 & 0.571 & \textbf{19.4} & 4.842 & 0.573 \\
    & $P_N(\h_\ell)$ & 0.065 & 0.134 & 0.208 & 278.5 & 7.453 & 0.263 \\ \bottomrule
    \end{tabular}
\end{table}

\begin{table}[t]
\centering
\caption{Held-out evaluation results for Qwen, OLMo, and LLaMA with direct PCA tangent spaces. Pass@$k$ and median rank are computed with the highest activating passing token, with Pass@$k$ aggregated over all sequences, and median rank averaged across the six suites.}
\label{tab:aggregate_pca_main}
\small 
\begin{tabular}{cccccccc}
\toprule
Model & Projection & P@1 $\uparrow$ & P@5 $\uparrow$ & P@10 $\uparrow$ & Mean med. rank $\downarrow$ & KL $\downarrow$ & Cos. $\uparrow$ \\
\midrule
& $\h_\ell$ & 0.195 & 0.356 & 0.407 & 461.8 & 10.429 & 0.405 \\
Qwen-2.5 7B & $P_{T}(\h_\ell)$ & \textbf{0.205} & \textbf{0.379} & \textbf{0.439} & \textbf{193} & \textbf{10.264} & \textbf{0.437} \\
& $P_N(\h_\ell)$ & 0.000 & 0.000 & 0.000 & 14437.6 & 17.148 & -0.053 \\ \midrule
& $\h_\ell$ & \textbf{0.204} & \textbf{0.395} & \textbf{0.507} & 251.3 & \textbf{7.299} & \textbf{0.485} \\
OLMo-2 7B & $P_T(\h_\ell)$ & 0.199 & 0.372 & 0.496 & \textbf{231.6} & 7.385 & 0.483 \\
& $P_N(\h_\ell)$ & 0.000 & 0.000 & 0.000 & 3532.3 & 9.374 & 0.173 \\ \midrule
& $\h_\ell$ & \textbf{0.305} & \textbf{0.480} & \textbf{0.604} & \textbf{23.9} & \textbf{4.719} & \textbf{0.606} \\
LLaMA-3.2 3B & $P_{T}(\h_\ell)$ & 0.301 & 0.447 & 0.519 & 28.7 & 5.299 & 0.576 \\
& $P_N(\h_\ell)$ & 0.145 & 0.222 & 0.233 & 576.3 & 7.514 & 0.290 \\
\bottomrule
\end{tabular}
\end{table}

The performance of this decomposition on OLMo is notably worse than Qwen and LLaMA. Given the results of Figure \ref{fig:tangent_decomp_by_layer}, where the excess overlap with the PCA tangent space is quite low, we hypothesize that our pushforward tangent approximation is poor on OLMo-2. OLMo obtains more similar results to Qwen and LLaMA with the PCA tangent space. These results, along with breakdowns by evaluation suite, can be found in Appendix \ref{sec:extended_results}.

\subsection{Layerwise decomposition of the tangent projected latents}
\begin{figure}
    \centering
    \includegraphics[width=\linewidth]{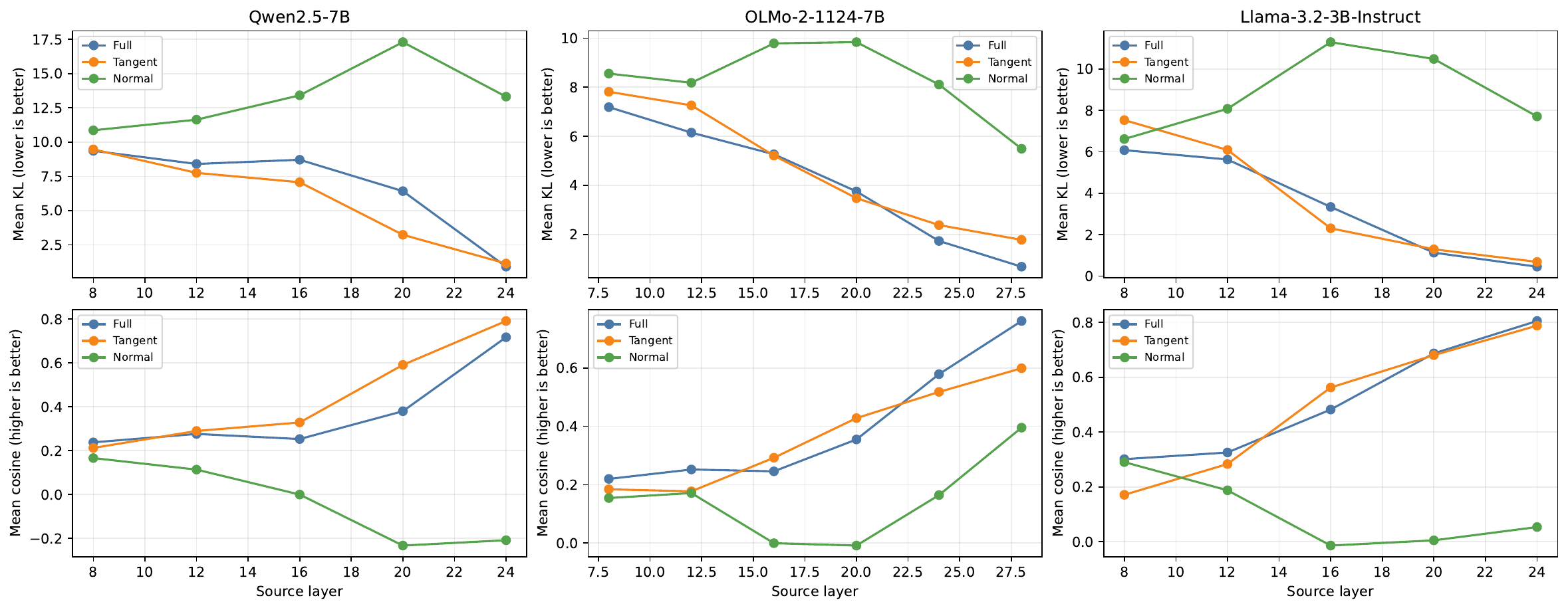}
    \caption{Layerwise average KL divergence (top) and cosine similarity (bottom) between the network output $f(\h_\ell)$ and the three linear approximations we study: $\BE[J]\h_\ell$, $\BE[J]P_T\h_\ell$, and $\BE[J]P_N\h_\ell$.}
    \label{fig:layerwise_kl_oracle}
\end{figure}

In addition to recording benchmark data, we tracked the KL divergence between $\BE[J]h_\ell$ and the actual output of the transformer, $h_L$, when projected into token space. This gives a particularly clean tool to examine the decomposition of hidden states into tangent and normal components independently of the studied benchmarks. We report the layerwise breakdown for the pushforward tangent spaces in Figure \ref{fig:layerwise_kl_oracle}. The tangent projection is most useful, for all three models, specifically in the mid-to-late layers most associated with the linear representation hypothesis. In Qwen, the denoising effect observed in Section \ref{sec:local_oracle} is visible in the way that this advantage is preserved in late layers of the network. 

\subsection{Exact local Jacobians}
To study the effect averaging has $\int_M \frac{\partial h_L}{\partial h_\ell}d\mu$ on the Jacobian lens, we additionally recorded results when, instead, the exact local Jacobian is used at each point on the activation manifold, shown in Table  \ref{tab:local_jacobian_results}. For ease of reference, we reproduce the results of Tables \ref{tab:aggregate_local_oracle}. 

\begin{table}
\centering
\caption{Local Jacobian results for Qwen, OLMo, and LLaMA, with expected Jacobian results for comparison.} \label{tab:local_jacobian_results}
\small 
\begin{tabular}{cccccccc}
\toprule
Model & Variant & P@1 $\uparrow$ & P@5 $\uparrow$ & P@10 $\uparrow$ & Mean med. rank $\downarrow$ & KL $\downarrow$ & Cos. $\uparrow$ \\
\midrule
& $\mathbb{E}[J]\h_\ell$ & 0.195 & 0.356 & 0.407 & 461.8 & 10.429 & 0.405 \\
& $\mathbb{E}[J]P_T(\h_\ell)$ & 0.267 & 0.437 & 0.546 & 137.0 & 9.286 & 0.470 \\
& $\mathbb{E}[J]P_N(\h_\ell)$ & 0.002 & 0.011 & 0.011 & 9609.4 & 15.801 & -0.098 \\
Qwen-2.5 7B & $J_{\h_\ell}\h_\ell$ & 0.295 & 0.397 & 0.414 & 466.2 & 6.762 & 0.373 \\
& $J_{\h_\ell} P_T(\h_\ell)$ & 0.236 & 0.367 & 0.391 & 616.2 & 5.735 & 0.443 \\
& $J_{\h_\ell}P_N(\h_\ell)$ & 0.012 & 0.048 & 0.085 & 2970.4 & 13.302 & -0.032 \\ \midrule
&  $\mathbb{E}[J]\h_\ell$ & 0.204 & 0.395 & 0.507 & 251.3 & 7.299 & 0.485 \\
& $\mathbb{E}[J]P_T(\h_\ell)$ & 0.187 & 0.363 & 0.419 & 169.5 & 7.801 & 0.422 \\
&  $\mathbb{E}[J]P_N(\h_\ell)$ & 0.142 & 0.277 & 0.352 & 528.1 & 7.899 & 0.366 \\
OLMo-2 7B & $J_{\h_\ell}\h_\ell$ & 0.286 & 0.359 & 0.393 & 460.9 & 4.132 & 0.402 \\
& $J_{\h_\ell}P_T(\h_\ell)$ & 0.290 & 0.318 & 0.349 & 590.1 & 4.657 & 0.367 \\
& $J_{\h_\ell}P_N(\h_\ell)$ & 0.041 & 0.086 & 0.101 & 871.0 & 8.328 & 0.146 \\ \midrule
&  $\mathbb{E}[J]\h_\ell$ & 0.305 & 0.480 & 0.604 & 23.9 & 4.719 & 0.606 \\
&  $\mathbb{E}[J]P_T(\h_\ell)$ & 0.320 & 0.473 & 0.571 & 19.4 & 4.842 & 0.573 \\
& $\mathbb{E}[J]P_N(\h_\ell)$ & 0.065 & 0.134 & 0.208 & 278.5 & 7.453 & 0.263 \\
LLaMA-3.2 3B & $J_{\h_\ell}\h_\ell$ & 0.238 & 0.430 & 0.483 & 187.3 & 3.325 & 0.520 \\
&  $J_{\h_\ell}P_T(\h_\ell)$ & 0.237 & 0.436 & 0.483 & 369.0 & 3.581 & 0.497 \\
& $J_{\h_\ell}P_N(\h_\ell)$ & 0.082 & 0.166 & 0.199 & 572.3 & 8.837 & 0.105 \\
\bottomrule
\end{tabular}
\end{table}

\subsection{Local projection results by evaluation suite}
We decompose the macroscopic results into their suite-by-suite decomposition in Table \ref{tab:suite_median_ranks}. Qwen shows that the tangent projection is a clean improvement on the full Jacobian across all suites, and LLaMA similarly shows very comparable performance between the tangent and full Jacobians. OLMo shows more mixed results. One confounding factor, which likely damages the performance of the exact local Jacobians $J_{\h_\ell}$, is that the Jacobian isn't determined purely by the network weights, but also by the specific contextual information, which adds noise to the signal.

\begin{table*}[t]
\centering
\caption{Median best ranks for the three models, divided by evaluation suite, on held-out data. Bolded are the best global mean and best local Jacobian approximation for each suite in each model family.}
\label{tab:suite_median_ranks}
\small
\begin{tabular}{clrrrrrr}
\toprule
Model & Projection & Multihop & Order-Ops & Multilingual & Typo & Poetry & Association \\ \midrule
&  $\mathbb{E}[J]\h_\ell$ & 5 & 35 & 6 & 65.5 & \textbf{1.5} & 2658 \\
& $\mathbb{E}[J]P_T(\h_\ell)$ & \textbf{3} & \textbf{6} & \textbf{5} & \textbf{30.5} & \textbf{1.5} & \textbf{776} \\
& $\mathbb{E}[J]P_N(\h_\ell)$ & 12723 & 12942 & 704 & 3923.5 & 6471 & 20893 \\
Qwen-2.5 7B & $J_{\h_\ell}\h_\ell$ & 45 & \textbf{1} & \textbf{53} & 194.5 & \textbf{1} & \textbf{2502.5} \\
& $J_{\h_\ell}P_T(\h_\ell)$ & \textbf{25} & \textbf{1} & 64 & \textbf{130} & \textbf{1} & 3476.5 \\
& $J_{\h_\ell}P_N(\h_\ell)$ & 4538 & 566 & 395 & 3417.5 & 673 & 8233 \\ \midrule 
& $\mathbb{E}[J]\h_\ell$ & \textbf{15} & 7 & 9 & 14.5 & \textbf{1} & 1461.5 \\
& $\mathbb{E}[J]P_T(\h_\ell)$ & 29 & 26 & \textbf{7} & \textbf{9} & \textbf{1} & \textbf{945} \\
& $\mathbb{E}[J]P_N(\h_\ell)$ & 19 & \textbf{3} & 48 & 165.5 & 2 & 2931 \\
OLMo-2 7B & $J_{\h_\ell}\h_\ell$ & \textbf{74} & \textbf{1} & \textbf{89} & \textbf{353} & \textbf{1} & \textbf{2247.5} \\
& $J_{\h_\ell}P_T(\h_\ell)$ & 179 & \textbf{1} & 247 & 725 & \textbf{1} & 2387.5 \\
& $J_{\h_\ell}P_N(\h_\ell)$ & 162 & 74 & 192 & 578 & 82 & 4138 \\ \midrule 
& $\mathbb{E}[J]\h_\ell$ & \textbf{16} & \textbf{7} & \textbf{3} & 2 & \textbf{1} & 114.5 \\
& $\mathbb{E}[J]P_T(\h_\ell)$ & 23 & 17 & \textbf{3} & \textbf{1} & \textbf{1} & \textbf{71.5} \\
& $\mathbb{E}[J]P_N(\h_\ell)$ & 104 & 16 & 41 & 539.5 & 47 & 923.5 \\
LLaMA-3.2 3B & $J_{\h_\ell}\h_\ell$ & \textbf{18} & 3 & \textbf{23} & 98.5 & \textbf{1} & \textbf{980.5} \\
& $J_{\h_\ell}P_T(\h_\ell)$ & \textbf{18} & \textbf{2} & 46 & \textbf{70.5} & \textbf{1} & 2076.5 \\
& $J_{\h_\ell}P_N(\h_\ell)$ & 553 & 395 & 28 & 959 & 6 & 1493 \\
\bottomrule
\end{tabular}
\end{table*}

\subsection{Exact local Jacobians with PCA tangent spaces}

We likewise recorded the local projection results with PCA approximated tangent spaces, and report them in Table \ref{tab:heldout_direct_pca_qwen25_7b_overall}. As above, we duplicate the projection of the global results here for easier comparison. 
\begin{table}[t]
\centering
\caption{Held-out evaluation results for Qwen, OLMo, and LLaMA with direct PCA tangent spaces, including both using the global approximator $\BE[J]$ and exact local Jacobians $J_{\h_\ell}$.}
\label{tab:heldout_direct_pca_qwen25_7b_overall}
\small
\begin{tabular}{ccrrrrrr}
\toprule
Model & Variant & P@1 & P@5 & P@10 & Mean med. rank & KL $\downarrow$ & Cos. $\uparrow$ \\
\midrule
& $\mathbb{E}[J]\h_\ell$ & 0.195 & 0.356 & 0.407 & 461.8 & 10.429 & 0.405 \\
& $\mathbb{E}[J]P_{T}(\h_\ell)$ & 0.205 & 0.379 & 0.439 & 193 & 10.264 & 0.437 \\
& $\mathbb{E}[J]P_N(\h_\ell)$ & 0.000 & 0.000 & 0.000 & 14437.6 & 17.148 & -0.053 \\
Qwen-2.5 7B & $J_{\h_\ell}\h_\ell$ & 0.295 & 0.397 & 0.414 & 466.2 & 6.762 & 0.373 \\
& $J_{\h_\ell}P_T(\h_\ell)$ & 0.229 & 0.326 & 0.379 & 1052.2 & 6.541 & 0.391 \\
& $J_{\h_\ell}P_N(\h_\ell)$ & 0.022 & 0.049 & 0.056 & 3095.7 & 13.801 & -0.040 \\ \midrule
& $\mathbb{E}[J]\h_\ell$ & 0.204 & 0.395 & 0.507 & 251.3 & 7.299 & 0.485 \\
& $\mathbb{E}[J]P_T(\h_\ell)$ & 0.199 & 0.372 & 0.496 & 231.6 & 7.385 & 0.483 \\
& $\mathbb{E}[J]P_N(\h_\ell)$ & 0.000 & 0.000 & 0.000 & 3532.3 & 9.374 & 0.173 \\
OLMo-2 7B & $J_{\h_\ell}\h_\ell$ & 0.286 & 0.359 & 0.393 & 460.9 & 4.132 & 0.402 \\
& $J_{\h_\ell}P_{T}(\h_\ell)$ & 0.281 & 0.350 & 0.384 & 440.2 & 4.014 & 0.423 \\
& $J_{\h_\ell}P_N(\h_\ell)$ & 0.000 & 0.033 & 0.058 & 2739.5 & 11.973 & -0.133 \\ \midrule
& $\mathbb{E}[J]\h_\ell$ & 0.305 & 0.480 & 0.604 & 23.9 & 4.719 & 0.606 \\
& $\mathbb{E}[J]P_{T}(\h_\ell)$ & 0.301 & 0.447 & 0.519 & 28.7 & 5.299 & 0.576 \\
& $\mathbb{E}[J]P_N(\h_\ell)$ & 0.145 & 0.222 & 0.233 & 576.3 & 7.514 & 0.290 \\
LLaMA-3.2 3B & $J_{\h_\ell}\h_\ell$ & 0.238 & 0.430 & 0.483 & 187.3 & 3.325 & 0.520 \\
& $J_{\h_\ell}P_{T}(\h_\ell)$ & 0.216 & 0.346 & 0.393 & 437.8 & 4.048 & 0.419 \\
& $J_{\h_\ell}P_N(\h_\ell)$ & 0.101 & 0.214 & 0.279 & 271.2 & 9.666 & 0.086 \\
\bottomrule
\end{tabular}
\end{table}

\subsection{Tangent- and normal-projected Jacobians by evaluation suite}
We give a breakdown of the results reported for each model family, divided by evaluation suite, in Tables \ref{tab:qwen7b-projected-lens-breakdown}--\ref{tab:llama3b-projected-lens-breakdown}. Unlike in the main body, here for overall median rank we report pooled median, rather than the unweighted average over suites.

\begin{table*}[t]
\centering
\small
\caption{Breakdown by evaluation suite for Qwen-2.5 7B of the $J$-lens tangent and normal projections}
\label{tab:qwen7b-projected-lens-breakdown}
\begin{tabular}{llrrrrr}
\toprule
Benchmark & Lens & Pass@1 & Pass@5 & Pass@10 & Mean KL & Median rank \\
\midrule
Multihop ($n=63$) & Logit lens & 12.7 & 22.2 & 25.4 & 9.10 & 196 \\
 & $\mathbb{E}[J]$ & 30.2 & 54.0 & 57.1 & 8.94 & 5 \\
 & $\mathbb{E}[J P_T]$ & 22.2 & 42.9 & 54.0 & 9.82 & 10 \\
 & $\mathbb{E}[J P_N]$ & 6.3 & 15.9 & 28.6 & 10.10 & 34 \\
\addlinespace
Order-ops ($n=25$) & Logit lens & 28.0 & 56.0 & 68.0 & 7.18 & 3 \\
 & $\mathbb{E}[J]$ & 16.0 & 24.0 & 32.0 & 12.96 & 35 \\
 & $\mathbb{E}[J P_T]$ & 20.0 & 40.0 & 52.0 & 13.46 & 7 \\
 & $\mathbb{E}[J P_N]$ & 0.0 & 0.0 & 4.0 & 13.89 & 269 \\
\addlinespace
Multilingual ($n=77$) & Logit lens & 27.3 & 48.1 & 51.9 & 10.30 & 8 \\
 & $\mathbb{E}[J]$ & 23.4 & 51.9 & 54.5 & 10.35 & 5 \\
 & $\mathbb{E}[J P_T]$ & 5.2 & 6.5 & 9.1 & 11.51 & 93 \\
 & $\mathbb{E}[J P_N]$ & 26.0 & 50.6 & 55.8 & 10.65 & 5 \\
\addlinespace
Typo ($n=66$) & Logit lens & 3.0 & 15.2 & 19.7 & 10.73 & 77 \\
 & $\mathbb{E}[J]$ & 1.5 & 3.0 & 10.6 & 10.44 & 80 \\
 & $\mathbb{E}[J P_T]$ & 0.0 & 1.5 & 4.5 & 9.48 & 223.5 \\
 & $\mathbb{E}[J P_N]$ & 1.5 & 10.6 & 16.7 & 12.14 & 81 \\
\addlinespace
Poetry ($n=68$) & Logit lens & 42.6 & 61.8 & 73.5 & 10.05 & 2 \\
 & $\mathbb{E}[J]$ & 50.0 & 80.9 & 83.8 & 10.64 & 1.5 \\
 & $\mathbb{E}[J P_T]$ & 35.3 & 57.4 & 64.7 & 12.83 & 3.5 \\
 & $\mathbb{E}[J P_N]$ & 41.2 & 72.1 & 79.4 & 10.85 & 2 \\
\addlinespace
Association ($n=72$) & Logit lens & 0.0 & 0.0 & 0.0 & 7.97 & 2{,}988.5 \\
 & $\mathbb{E}[J]$ & 0.0 & 0.0 & 0.0 & 8.51 & 2{,}233 \\
 & $\mathbb{E}[J P_T]$ & 0.0 & 0.0 & 0.0 & 10.84 & 1{,}520 \\
 & $\mathbb{E}[J P_N]$ & 1.4 & 1.4 & 2.8 & 8.52 & 3{,}658 \\
\midrule
Overall ($n=371$) & Logit lens & 18.1 & 31.5 & 36.7 & 9.46 & 39 \\
 & $\mathbb{E}[J]$ & 20.5 & 36.9 & 40.4 & 10.00 & 24 \\
 & $\mathbb{E}[J P_T]$ & 12.7 & 22.1 & 27.2 & 11.10 & 84 \\
 & $\mathbb{E}[J P_N]$ & 14.6 & 28.6 & 34.8 & 10.66 & 46 \\
\bottomrule
\end{tabular}
\end{table*}

\begin{table*}[t]
\centering
\small
\caption{Breakdown by evaluation suite for OLMo-2 7B of the $J$-lens, as well as its tangent and normal projections.}
\label{tab:olmo7b-projected-lens-breakdown}
\begin{tabular}{llrrrrr}
\toprule
Benchmark & Lens & Pass@1 & Pass@5 & Pass@10 & Mean KL & Median rank \\
\midrule
Multihop ($n=63$) & Logit lens & 12.7 & 30.2 & 34.9 & 6.03 & 38 \\
 & $\mathbb{E}[J]$ & 14.3 & 28.6 & 36.5 & 7.91 & 28 \\
 & $\mathbb{E}[J P_T]$ & 6.3 & 15.9 & 22.2 & 7.55 & 51 \\
 & $\mathbb{E}[J P_N]$ & 15.9 & 27.0 & 36.5 & 8.60 & 25 \\
\addlinespace
Order-ops ($n=25$) & Logit lens & 68.0 & 80.0 & 80.0 & 5.82 & 1 \\
 & $\mathbb{E}[J]$ & 0.0 & 12.0 & 44.0 & 9.32 & 11 \\
 & $\mathbb{E}[J P_T]$ & 0.0 & 0.0 & 4.0 & 8.99 & 38 \\
 & $\mathbb{E}[J P_N]$ & 0.0 & 12.0 & 28.0 & 9.81 & 18 \\
\addlinespace
Multilingual ($n=77$) & Logit lens & 27.3 & 58.4 & 70.1 & 6.86 & 3 \\
 & $\mathbb{E}[J]$ & 23.4 & 37.7 & 44.2 & 6.42 & 20 \\
 & $\mathbb{E}[J P_T]$ & 3.9 & 13.0 & 20.8 & 6.67 & 78 \\
 & $\mathbb{E}[J P_N]$ & 24.7 & 33.8 & 44.2 & 6.96 & 20 \\
\addlinespace
Typo ($n=66$) & Logit lens & 4.5 & 19.7 & 30.3 & 6.25 & 31 \\
 & $\mathbb{E}[J]$ & 22.7 & 57.6 & 72.7 & 7.79 & 3 \\
 & $\mathbb{E}[J P_T]$ & 0.0 & 1.5 & 3.0 & 6.61 & 964 \\
 & $\mathbb{E}[J P_N]$ & 28.8 & 69.7 & 77.3 & 8.89 & 2 \\
\addlinespace
Poetry ($n=68$) & Logit lens & 79.4 & 92.6 & 94.1 & 5.61 & 1 \\
 & $\mathbb{E}[J]$ & 60.3 & 83.8 & 88.2 & 9.83 & 1 \\
 & $\mathbb{E}[J P_T]$ & 23.5 & 48.5 & 54.4 & 10.39 & 6.5 \\
 & $\mathbb{E}[J P_N]$ & 58.8 & 83.8 & 88.2 & 10.10 & 1 \\
\addlinespace
Association ($n=72$) & Logit lens & 1.4 & 6.9 & 9.7 & 5.56 & 526.5 \\
 & $\mathbb{E}[J]$ & 0.0 & 0.0 & 1.4 & 7.54 & 1{,}537.5 \\
 & $\mathbb{E}[J P_T]$ & 0.0 & 0.0 & 0.0 & 8.43 & 4{,}552 \\
 & $\mathbb{E}[J P_N]$ & 0.0 & 1.4 & 1.4 & 7.54 & 1{,}023.5 \\
\midrule
Overall ($n=371$) & Logit lens & 28.0 & 44.5 & 50.4 & 6.06 & 10 \\
 & $\mathbb{E}[J]$ & 22.4 & 39.1 & 47.7 & 7.96 & 14 \\
 & $\mathbb{E}[J P_T]$ & 6.2 & 14.6 & 18.9 & 7.99 & 125 \\
 & $\mathbb{E}[J P_N]$ & 23.7 & 40.4 & 47.4 & 8.46 & 13 \\
\bottomrule
\end{tabular}
\end{table*}

\begin{table*}[t]
\centering
\small
\caption{Breakdown by evaluation suite for LLaMA-3.2 3B of the Jacobian, as well as its intrinsic tangent and extrinsic normal projections. }
\label{tab:llama3b-projected-lens-breakdown}
\begin{tabular}{llrrrrr}
\toprule
Benchmark & Lens & Pass@1 & Pass@5 & Pass@10 & Mean KL & Median rank \\
\midrule
Multihop ($n=63$) & Logit lens & 7.9 & 17.5 & 23.8 & 5.82 & 69 \\
 & $\mathbb{E}[J]$ & 14.3 & 33.3 & 36.5 & 4.32 & 23 \\
 & $\mathbb{E}[J P_T]$ & 6.3 & 19.0 & 28.6 & 4.48 & 32 \\
 & $\mathbb{E}[J P_N]$ & 14.3 & 31.7 & 38.1 & 5.36 & 21 \\
\addlinespace
Order-ops ($n=25$) & Logit lens & 8.0 & 36.0 & 36.0 & 6.73 & 18 \\
 & $\mathbb{E}[J]$ & 0.0 & 32.0 & 64.0 & 4.36 & 8 \\
 & $\mathbb{E}[J P_T]$ & 8.0 & 36.0 & 60.0 & 4.33 & 8 \\
 & $\mathbb{E}[J P_N]$ & 0.0 & 12.0 & 28.0 & 5.01 & 17 \\
\addlinespace
Multilingual ($n=77$) & Logit lens & 40.3 & 53.2 & 63.6 & 7.10 & 3 \\
 & $\mathbb{E}[J]$ & 42.9 & 66.2 & 75.3 & 6.50 & 2 \\
 & $\mathbb{E}[J P_T]$ & 11.7 & 37.7 & 46.8 & 6.77 & 13 \\
 & $\mathbb{E}[J P_N]$ & 44.2 & 63.6 & 71.4 & 6.93 & 2 \\
\addlinespace
Typo ($n=66$) & Logit lens & 60.6 & 81.8 & 87.9 & 6.13 & 1 \\
 & $\mathbb{E}[J]$ & 47.0 & 65.2 & 69.7 & 4.56 & 2 \\
 & $\mathbb{E}[J P_T]$ & 0.0 & 0.0 & 4.5 & 4.40 & 110 \\
 & $\mathbb{E}[J P_N]$ & 57.6 & 68.2 & 71.2 & 6.15 & 1 \\
\addlinespace
Poetry ($n=68$) & Logit lens & 80.9 & 97.1 & 97.1 & 5.63 & 1 \\
 & $\mathbb{E}[J]$ & 72.1 & 89.7 & 97.1 & 4.66 & 1 \\
 & $\mathbb{E}[J P_T]$ & 55.9 & 82.4 & 94.1 & 6.00 & 1 \\
 & $\mathbb{E}[J P_N]$ & 67.6 & 89.7 & 97.1 & 4.91 & 1 \\
\addlinespace
Association ($n=72$) & Logit lens & 1.4 & 5.6 & 9.7 & 6.11 & 684.5 \\
 & $\mathbb{E}[J]$ & 4.2 & 6.9 & 16.7 & 3.70 & 87.5 \\
 & $\mathbb{E}[J P_T]$ & 0.0 & 0.0 & 0.0 & 4.57 & 389.5 \\
 & $\mathbb{E}[J P_N]$ & 4.2 & 9.7 & 15.3 & 4.80 & 155 \\
\midrule
Overall ($n=371$) & Logit lens & 36.1 & 49.9 & 55.0 & 6.23 & 6 \\
 & $\mathbb{E}[J]$ & 33.7 & 50.9 & 59.6 & 4.76 & 5 \\
 & $\mathbb{E}[J P_T]$ & 14.3 & 28.6 & 36.7 & 5.23 & 34 \\
 & $\mathbb{E}[J P_N]$ & 35.0 & 49.9 & 56.6 & 5.61 & 6 \\
\bottomrule
\end{tabular}
\end{table*}

\subsection{Layerwise coherence and Frobenius statistics for projected lenses}
In Figure \ref{fig:relative_coherence}, we summarized the cancellation of the tangent projected Jacobians in terms of relative coherence. In Tables \ref{tab:qwen_coherence}--\ref{tab:llama_coherence}, we provide extended results from this study, including tracking the local Frobenius norm change, which tracks precisely with the relative rank of the approximated tangent space, and the Frobenius norm of the global objects, which substantially decreases (local projections lose coherence) as layer index increases.

\begin{table}[t]
\centering
\small
\caption{Rank, directional-energy, and coherence decomposition for Qwen-2.5 7B. Relative rank is the norm ratio $\sqrt{\text{rank}/d}$ weighted by per-sample full-Jacobian energy; relative mean Frobenius norm is the ratio of local Frobenius norms $\sqrt{\mathbb{E}\|JP_T\|_F^2/\mathbb{E}\|J\|_F^2}$; Relative Frobenius norm is $\|\mathbb{E}[JP_T]\|_F/\|\mathbb{E}[J]\|_F$; and relative coherence is the ratio of the tangent to total coherence, where coherence is $\frac{||\frac1n\sum J_xP_x||^2_F}{\frac1n \sum ||J_xP_x||^2_F}$, where $P_x = P_{T_x}$ for the tangent projection, and $I$ for the full map.}\label{tab:qwen_coherence}
\begin{tabular}{ccccc}
\toprule
Layer & Relative rank & Relative Mean Frob. Norm & Relative Frob. Norm &  Relative Coherence \\
\midrule
4 & 0.407 & 0.421 & 0.336 & 0.637 \\
8 &  0.400 & 0.448 & 0.378 &  0.715 \\
12 & 0.375 & 0.401 & 0.376 &  0.880 \\
16  & 0.363 & 0.427 & 0.355 & 0.690 \\
20  & 0.337 & 0.511 & 0.296 & 0.334 \\
24  & 0.304 & 0.464 & 0.164 & 0.126 \\
26  & 0.298 & 0.322 & 0.119 & 0.136 \\
\bottomrule
\end{tabular}
\end{table}

\begin{table}[t]
\centering
\small
\caption{Rank, directional-energy, and coherence decomposition for OLMo-2 7B. Relative rank is the norm ratio $\sqrt{\text{rank}/d}$ weighted by per-sample full-Jacobian energy; relative mean Frobenius norm is the ratio of local Frobenius norms $\sqrt{\mathbb{E}\|JP_T\|_F^2/\mathbb{E}\|J\|_F^2}$; Relative Frobenius norm is $\|\mathbb{E}[JP_T]\|_F/\|\mathbb{E}[J]\|_F$; and relative coherence is the ratio of the tangent to total coherence, where coherence is $\frac{||\frac1n\sum J_xP_x||^2_F}{\frac1n \sum ||J_xP_x||^2_F}$, where $P_x = P_{T_x}$ for the tangent projection, and $I$ for the full map.}\label{tab:olmo_coherence}
\begin{tabular}{ccccc}
\toprule
Layer & Relative rank & Relative Mean Frob. Norm & Relative Frob. Norm &  Relative Coherence \\
\midrule
4 & 0.319 & 0.352 & 0.271 & 0.590 \\
8 & 0.340 & 0.399 & 0.335 & 0.706 \\
12 & 0.294 & 0.368 & 0.305 & 0.689 \\
16 & 0.233 & 0.294 & 0.257 & 0.767 \\
20 & 0.200 & 0.228 & 0.197 & 0.747 \\
24 & 0.199 & 0.215 & 0.167 & 0.604 \\
28 & 0.182 & 0.191 & 0.127 & 0.443 \\
30 & 0.161 & 0.154 & 0.104 & 0.461 \\
\bottomrule
\end{tabular}
\end{table}

\begin{table}[t]
\centering
\small
\caption{Rank, directional-energy, and coherence decomposition for LLaMA-3.2 3B. Relative rank is the norm ratio $\sqrt{\text{rank}/d}$ weighted by per-sample full-Jacobian energy; relative mean Frobenius norm is the ratio of local Frobenius norms $\sqrt{\mathbb{E}\|JP_T\|_F^2/\mathbb{E}\|J\|_F^2}$; Relative Frobenius norm is $\|\mathbb{E}[JP_T]\|_F/\|\mathbb{E}[J]\|_F$; and relative coherence is the ratio of the tangent to total coherence, where coherence is $\frac{||\frac1n\sum J_xP_x||^2_F}{\frac1n \sum ||J_xP_x||^2_F}$, where $P_x = P_{T_x}$ for the tangent projection, and $I$ for the full map.}\label{tab:llama_coherence}
\begin{tabular}{ccccc}
\toprule
Layer & Relative rank & Relative Mean Frob. Norm & Relative Frob. Norm &  Relative Coherence \\
\midrule
4 & 0.358 & 0.402 & 0.326 & 0.660 \\
8 & 0.390 & 0.481 & 0.370 & 0.593 \\
12 & 0.369 & 0.497 & 0.351 & 0.498 \\
16 & 0.314 & 0.395 & 0.210 & 0.282 \\
20 & 0.281 & 0.330 & 0.132 & 0.160 \\
24 & 0.277 & 0.297 & 0.105 & 0.126 \\
26 & 0.285 & 0.284 & 0.105 & 0.136 \\
\bottomrule
\end{tabular}
\end{table}

\subsection{Alternative norms on $H_\ell$}

We briefly investigated two alternative norms on $H_\ell$, the Mahalanobis geometry and the Fisher metric. To construct the projection, given an orthonormal (Euclidean) basis for the tangent space $B_\ell$ and a positive-definite metric $G$, 
\[
P_{T,G} = B_\ell(B_\ell^\top G B_\ell)^{-1}B_\ell^\top G.
\]

The Mahalanobis geometry whitens activations, emphasizing distance in rarely varying directions. In this case, $G \propto (\operatorname{Cov}(\h_\ell)+\lambda c\operatorname{Id})^{-1}$, where $c = \frac{\operatorname{Tr}\operatorname{Cov}(\h_\ell)}{d}$ is a normalizing constant, and $\lambda$ is the ridge regression factor. We estimated this using 3200 activations per layer of LLaMA-3.2 3B, and a ridge regression of 0.01. We found (Tables \ref{tab:llama-intrinsic-local-oracle} and \ref{tab:llama-intrinsic-projected-jacobian-lens}) that whitening harms tangent-projection performance and strongly improves normal-projection performance in both experiments. We selected 0.01 after performing a sweep, and selecting it as the factor which most improved normal performance. 

The local Fisher metric at $\h_\ell$ is
\[
  G(\h_\ell) = J_{\h_\ell\to \h_L}^{\top} \left(\operatorname{diag}(p)-pp^\top\right)J_{\h_\ell\to \h_L},
\]
where $p$ are the token probabilities (i.e. softmax of logits). This norm reflects directly how small variation in the hidden state affects output probability distributions. We also calculated an averaged Fisher metric, which averages the calculated norm on a generated token across 3200 contexts per layer. By sampling and averaging at the sampled token, we obtain an unbiased Monte-Carlo estimate of the local Fisher metric averaged across all contexts. In Tables \ref{tab:llama-intrinsic-local-oracle} and \ref{tab:llama-intrinsic-projected-jacobian-lens}, we show that both the local and averaged Fisher metrics conform to the Euclidean metric results.

\begin{table}[t]
\centering
\small
\caption{Local-oracle results for Llama-3.2 3B. Each of the six benchmark suites receives equal weight; mean median rank is the mean of suite medians. The local Fisher row uses a fixed top-128 support and $\lambda=10^{-3}$.}
\label{tab:llama-intrinsic-local-oracle}
\begin{tabular}{cccccccc}
\toprule
Inner product & Variant & P@1 $\uparrow$ & P@5 $\uparrow$ & P@10 $\uparrow$ & Mean med. rank $\downarrow$ & KL $\downarrow$ & Cos. $\uparrow$ \\
\midrule
 & $\mathbb{E}[J]\h_\ell$ & 0.305 & 0.480 & 0.604 & 23.9 & 4.719 & 0.606 \\
 & $J_{\h_\ell}\h_\ell$ & 0.238 & 0.430 & 0.483 & 187.3 & 3.325 & 0.520 \\
\midrule
Euclidean & $\mathbb{E}[J]P_T(\h_\ell)$ & 0.320 & 0.473 & 0.571 & 19.4 & 4.842 & 0.573 \\
 & $\mathbb{E}[J]P_N(\h_\ell)$ & 0.065 & 0.134 & 0.208 & 278.5 & 7.453 & 0.263 \\
 & $J_{\h_\ell}P_T(\h_\ell)$ & 0.237 & 0.436 & 0.483 & 369.0 & 3.581 & 0.497 \\
 & $J_{\h_\ell}P_N(\h_\ell)$ & 0.082 & 0.166 & 0.199 & 572.3 & 8.837 & 0.105 \\
\midrule
Mahalanobis & $\mathbb{E}[J]P_T(\h_\ell)$ & 0.298 & 0.421 & 0.524 & 24.2 & 5.983 & 0.485 \\
 & $\mathbb{E}[J]P_N(\h_\ell)$ & 0.240 & 0.369 & 0.450 & 86.7 & 5.769 & 0.421 \\
 & $J_{\h_\ell}P_T(\h_\ell)$ & 0.163 & 0.382 & 0.433 & 361.3 & 4.085 & 0.393 \\
 & $J_{\h_\ell}P_N(\h_\ell)$ & 0.072 & 0.207 & 0.285 & 214.5 & 7.935 & 0.209 \\
\midrule
Avg. output Fisher & $\mathbb{E}[J]P_T(\h_\ell)$ & 0.321 & 0.487 & 0.554 & 28.0 & 5.116 & 0.486 \\
 & $\mathbb{E}[J]P_N(\h_\ell)$ & 0.002 & 0.028 & 0.036 & 3409.5 & 11.264 & -0.018 \\
 & $J_{\h_\ell}P_T(\h_\ell)$ & 0.245 & 0.428 & 0.479 & 229.3 & 4.375 & 0.408 \\
 & $J_{\h_\ell}P_N(\h_\ell)$ & 0.007 & 0.036 & 0.057 & 3199.5 & 11.612 & -0.055 \\
\midrule
Local output Fisher & $\mathbb{E}[J]P_T(\h_\ell)$ & 0.269 & 0.457 & 0.560 & 18.4 & 5.127 & 0.504 \\
 & $\mathbb{E}[J]P_N(\h_\ell)$ & 0.133 & 0.244 & 0.328 & 164.6 & 7.117 & 0.280 \\
 & $J_{\h_\ell}P_T(\h_\ell)$ & 0.220 & 0.394 & 0.449 & 295.9 & 4.509 & 0.399 \\
 & $J_{\h_\ell}P_N(\h_\ell)$ & 0.014 & 0.048 & 0.083 & 698.9 & 7.621 & 0.195 \\
\bottomrule
\end{tabular}
\end{table}

\begin{table}[t]
\centering
\small
\caption{Tangent- and normal-projected Jacobian lenses for LLaMA-3.2 3B Euclidean, Mahalanobis, and Fisher metrics. Each suite receives equal weight; mean median rank is the mean of suite medians.}
\label{tab:llama-intrinsic-projected-jacobian-lens}
\begin{tabular}{cccccccc}
\toprule
Inner product & Variant & P@1 $\uparrow$ & P@5 $\uparrow$ & P@10 $\uparrow$ & Mean med. rank $\downarrow$ & KL $\downarrow$ & Cos. $\uparrow$ \\
\midrule
 & $\mathbb{E}[J]\h_\ell$ & 0.301 & 0.489 & 0.599 & 20.6 & 4.684 & 0.599 \\
\midrule
Euclidean & $\mathbb{E}[JP_T]\h_\ell$ & 0.139 & 0.292 & 0.392 & 92.8 & 5.087 & 0.538 \\
 & $\mathbb{E}[JP_N]\h_\ell$ & 0.311 & 0.458 & 0.538 & 32.7 & 5.530 & 0.525 \\
\midrule
Mahalanobis & $\mathbb{E}[JP_T]\h_\ell$ & 0.033 & 0.114 & 0.158 & 265.8 & 8.189 & 0.336 \\
 & $\mathbb{E}[JP_N]\h_\ell$ & 0.327 & 0.546 & 0.618 & 22.8 & 4.791 & 0.576 \\
\midrule
Avg. output Fisher & $\mathbb{E}[JP_T]\h_\ell$ & 0.173 & 0.327 & 0.387 & 125.0 & 4.970 & 0.505 \\
 & $\mathbb{E}[JP_N]\h_\ell$ & 0.269 & 0.370 & 0.417 & 135.8 & 9.556 & 0.194 \\
\midrule
Local output Fisher & $\mathbb{E}[JP_T]\h_\ell$ & 0.215 & 0.343 & 0.397 & 110.5 & 4.869 & 0.539 \\
 & $\mathbb{E}[JP_N]\h_\ell$ & 0.316 & 0.444 & 0.503 & 49.0 & 6.703 & 0.439 \\
\bottomrule
\end{tabular}
\end{table}

\subsection{Alternative centers}
We briefly considered two alternative dataset centers. While 0 is architecturally specified by RMSNorm, we also studied the dataset mean activation $\mu_\ell$ and a fitted cone apex $c_\ell$. The latter is calculated by using the pushforward norm to calculate the tangent space across a wide variety of hidden states $\h_i$, and using least squares to find the point $c_\ell$ which is closest to the set of spans $\{\h_i + T_{\h_i}\M \}$. We show results in Table \ref{tab:llama-center-comparison}, where 0 mildly outperforms the alternatives.

\begin{table}
\centering 
\caption{Local-oracle results for LLaMA-3.2 3B, recentered to the origin, the dataset mean $\mu_\ell$, and a fitted cone apex $c_\ell$.}
\label{tab:llama-center-comparison}
\small 
\begin{tabular}{cccccccc}
\toprule
Center & Variant & P@1 $\uparrow$ & P@5 $\uparrow$ & P@10 $\uparrow$ & Mean med. rank $\downarrow$ & KL $\downarrow$ & Cos. $\uparrow$ \\
\midrule
 & $\mathbb{E}[J]\h_\ell$ & 0.302 & 0.478 & 0.601 & 22.0 & 4.760 & 0.605 \\
0 & $\mathbb{E}[J]P_{T_{\h_\ell}}\h_\ell$ & 0.326 & 0.484 & 0.580 & 19.7 & 4.875 & 0.571 \\
 & $\mathbb{E}[J]P_{N_{\h_\ell}}\h_\ell$ & 0.047 & 0.122 & 0.199 & 311.2 & 7.524 & 0.262 \\
\midrule
 & $\mathbb{E}[J](\h_\ell-\mu_\ell)$ & 0.264 & 0.376 & 0.447 & 61.0 & 6.174 & 0.488 \\
$\mu_\ell$ & $\mathbb{E}[J]P_{T_{\h_\ell}}(\h_\ell-\mu_\ell)$ & 0.299 & 0.453 & 0.564 & 20.1 & 5.813 & 0.520 \\
 & $\mathbb{E}[J]P_{N_{\h_\ell}}(\h_\ell-\mu_\ell)$ & 0.029 & 0.097 & 0.104 & 1162.7 & 8.985 & 0.131 \\
\midrule
 & $\mathbb{E}[J](\h_\ell- c_\ell)$ & 0.283 & 0.438 & 0.517 & 29.0 & 5.328 & 0.569 \\
$c_\ell$ & $\mathbb{E}[J]P_{T_{\h_\ell}}(\h_\ell- c_\ell)$ & 0.299 & 0.464 & 0.522 & 23.7 & 5.397 & 0.540 \\
 & $\mathbb{E}[J]P_{N_{\h_\ell}}(\h_\ell-c_\ell)$ & 0.083 & 0.156 & 0.214 & 524.3 & 7.490 & 0.266 \\
\bottomrule
\end{tabular}
\end{table}